\documentclass[sigconf]{acmart}
\AtBeginDocument{%
  }

\copyrightyear{2026}
\acmYear{2026}
\setcopyright{cc}
\setcctype{by}
\acmConference[WWW '26]{Proceedings of the ACM Web Conference 2026}{April 13--17, 2026}{Dubai, United Arab Emirates}
\acmBooktitle{Proceedings of the ACM Web Conference 2026 (WWW '26), April 13--17, 2026, Dubai, United Arab Emirates}
\acmPrice{}
\acmDOI{10.1145/3774904.3792237}
\acmISBN{979-8-4007-2307-0/2026/04}
\usepackage{subfigure}
\usepackage{multirow}
\usepackage{amsmath}
\usepackage{mathtools}
\usepackage{amsthm}
\usepackage{algorithm}
\usepackage{algorithmic}
\usepackage{wrapfig}
\usepackage[normalem]{ulem}
\useunder{\uline}{\ul}{}

\theoremstyle{plain}

\theoremstyle{definition}

\theoremstyle{remark}

\begin{document}

\title{Simple-Sampling and Hard-Mixup with Prototypes to Rebalance Contrastive Learning for Text Classification}


\author{Mengyu Li}
\authornote{Key Laboratory of Symbolic Computation and Knowledge Engineering of the Ministry of Education}
\authornote{Equal Contribution}
\author{Yonghao Liu}
\authornotemark[1]
\authornotemark[2]
\affiliation{
  \institution{College of Computer Science and Technology, Jilin University}
  \city{Changchun}
  \country{China}
}
\email{mengyul21@mails.jlu.edu.cn}
\email{yonghao20@mails.jlu.edu.cn}

\author{Fausto Giunchiglia}
\affiliation{%
  \institution{Department of Information Engineering and Computer Science, University of Trento}
  \city{Trento}
  \country{Italy}
}
\email{fausto.giunchiglia@unitn.it}

\author{Ximing Li}
\authornotemark[1]
\affiliation{%
  \institution{College of Computer Science and Technology, Jilin University}
  \city{Changchun}
  \country{China}
}
\email{liximing86@gmail.com}

\author{Xiaoyue Feng}
\authornotemark[1]
\authornote{Corresponding Author}
\affiliation{%
  \institution{College of Computer Science and Technology, Jilin University}
  \city{Changchun}
  \country{China}
}
\email{fengxy@jlu.edu.cn}

\author{Renchu Guan}
\authornotemark[1]
\authornotemark[3]
\affiliation{%
  \institution{College of Computer Science and Technology, Jilin University}
  \city{Changchun}
  \country{China}
}
\email{guanrenchu@jlu.edu.cn}

\renewcommand{\shortauthors}{Mengyu Li et al.}

\begin{abstract}
Text classification is a crucial and fundamental task in web content mining. Compared with the previous learning paradigm of pre-training and fine-tuning by cross entropy loss, the recently proposed supervised contrastive learning approach has received tremendous attention due to its powerful feature learning capability and robustness. Although several studies have incorporated this technique for text classification, some limitations remain. First, many text datasets are imbalanced, and the learning mechanism of supervised contrastive learning is sensitive to data imbalance, which may harm the model's performance. Moreover, these models leverage separate classification branches with cross entropy and supervised contrastive learning branches without explicit mutual guidance. To this end, we propose a novel model named SharpReCL for imbalanced text classification tasks. First, we obtain the prototype vector of each class in the balanced classification branch to act as a representation of each class. Then, by further explicitly leveraging the prototype vectors, we construct a proper and sufficient target sample set with the same size for each class to perform the supervised contrastive learning procedure. The empirical results show the effectiveness of our model, which even outperforms popular large language models across several datasets. Our code is available \href{https://github.com/KEAML-JLU/SharpReCL}{\textcolor{red}{here}}.
\end{abstract}



\maketitle

\section{Introduction}
Text classification (TC), as one of the fundamental tasks in the field of web content mining, has attracted much attention from academia and industry and is widely used in various fields \cite{liu2024improved}, such as question answering \cite{liulocal} and topic labeling tasks \cite{antypas2022twitter}. In TC processes, the ability of the model to extract textual features from raw text is crucial to the model performance \cite{liu2021deep, liu2025boosting, liu2025improved}. Text data in the form of symbolic sequences are extremely different from image data, and their discrete nature makes feature extraction and information utilization time-consuming and inefficient. 

Owing to the remarkable performance of large-scale pre-trained models, the learning paradigm of pre-training and fine-tuning has emerged as the new standard across various fields. In web content classification tasks, numerous models adhere to this paradigm and are fine-tuned under the guidance of cross-entropy (CE) loss to adapt to downstream tasks \cite{devlin2018bert,liu2019roberta}. Despite achieving state-of-the-art results on many datasets, CE loss exhibits certain limitations: a number of studies \cite{liu2019roberta,cao2019learning} have demonstrated that CE results in reduced generalization capabilities and lacks robustness to data noise \cite{sukhbaatar2014training}. Furthermore, CE exhibits instability across different runs when employed for fine-tuning in classification tasks, potentially impeding the model from attaining optimal performance \cite{dodge2020fine}. 

Recently, the emerging contrastive learning (CL) has showcased impressive representation learning capabilities and has been extensively applied in both computer vision (CV) and natural language processing (NLP) domains \cite{chen2020simple,he2020momentum,gao2021simcse}. Notably, supervised contrastive learning (SCL) \cite{khosla2020supervised}, a supervised variant of CL, aims to incorporate label information to extend the unsupervised InfoNCE loss to a supervised contrastive loss. This approach brings the representations of samples belonging to the same class closer together while pushing those of different classes apart. Some work \cite{hendrycks2019benchmarking} has demonstrated that SCL is less sensitive to hyperparameters in optimizers and data augmentation, exhibiting enhanced stability. Consequently, several studies \cite{gunel2020supervised,chen2022dual} have attempted to merge the learning methods of SCL and cross-entropy loss for fine-tuning PLMs, 
achieving considerable success.


\begin{figure}[ht]
    \centering
    \includegraphics[width=0.49\textwidth]{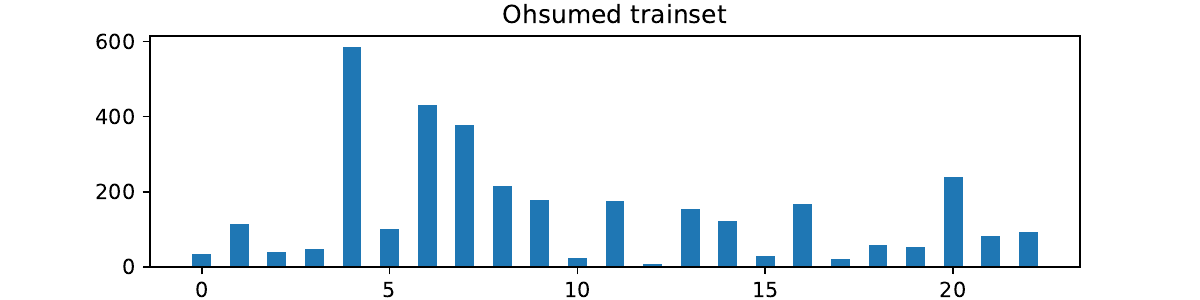}
    \caption{Label distribution of the Ohsumed trainset. The horizontal coordinate denotes the category, and the vertical coordinate denotes the corresponding frequency.}
    \label{imbalance}
\end{figure}

Despite the fruitful success of previous SCL-based models in TC tasks, there are several limitations that hinder the performance of these models. First, they all assume that the models are trained on large-scale, sufficient and balanced datasets, where each class has adequate and equal training samples \cite{zhu2022balanced}. However, in real-world application scenarios, the label distribution of data is often imbalanced. Especially when there are many classes in the dataset, many of which (minority class) have few samples, and a handful (majority class) have numerous samples. 
As shown in Fig. \ref{imbalance}, existing TC datasets such as the Ohsumed dataset \cite{yao2019graph} are often class-imbalanced. When deep models are trained with such datasets, they may yield suboptimal results and even suffer from overfitting problems, especially in minority classes \cite{zhang2021deep,yang2022multi}. Meanwhile, the manner of SCL in selecting positive and negative pairs also makes it more sensitive to data imbalance. 
Specifically, we suppose that the batch is $B$ and the imbalance ratio of the batch is $\text{ir}=\frac{\pi_\text{max}}{\pi_
\text{min}}$, where $\pi_\text{max}$ and $\pi_\text{min}$ are the frequencies of the majority class and the minority class in the original dataset, respectively. Since each instance should select samples from the same class to form the positive pairs and from other classes to form the negative pairs, 
the imbalance ratio can be further written as $\text{ir}=\frac{({|B|\pi_\text{max}})^2}{({|B|\pi_\text{min}})^2}=(\frac{{\pi_\text{max}}}{{\pi_\text{min}}})^2$ (the detailed analysis is shown in the ``Preliminaries'' section). That means when performing SCL on imbalanced datasets, the imbalance ratio is quadratic, and the imbalanced issue becomes more severe. Moreover, when the class imbalance problem exists in the dataset, samples may not encounter the proper positive or negative samples 
within a batch in the SCL paradigm \cite{song2022supervised}. That is, the majority class samples may lack sufficient negative samples and the minority class samples 
can also lack adequate positive samples. 
 Therefore, obtaining excellent model performance when performing SCL for imbalanced TC tasks is a great challenge, and there has been little effort to explore it. 


Second, previous models \cite{gunel2020supervised,suresh2021not} commonly utilize labeled data in two different branches, \textit{i.e.}, learning separately by following the paradigm of CE and of SCL 
and then obtaining the final training loss by a weighted average of the corresponding losses. However, such approaches only superficially combine CE and SCL, without any component or architecture linking CE and SCL to make them carry out further display interactions. It is also worth noting that the labeled data utilized for CE and SCL are the same, and these two modules both aim to achieve the effect of aggregating samples from the same class while separating samples from different classes. Therefore, further extracting effective information from these two supervised learning modules in an interactive manner to guide model training remains a challenge. 

To address the above issues, we propose SharpReCL, meaning \underline{S}imple-Sampling and \underline{Har}d-Mixup with \underline{P}rototypes to \underline{Re}balance \underline{C}ontrastive \underline{L}earning for TC. To address the problem of lack of mutual guidance between the two learning branches, we leverage the learned weight of classification branch to compute class prototype vectors, which serves as a bridge connecting to the SCL branch, where they act as anchor data and guide the generation of hard samples. 
Specifically, we replenish these prototype vectors as samples of the corresponding classes for each training batch, which ensures that every class can be sampled at least once, thus avoiding the situation where samples from the minority classes are not sampled. 
Moreover, for the class-imbalanced problem, in addition to resampling the original samples, we first identify those hard-to-distinguish sample embeddings of each class from the original data according to the similarity calculated with the prototype vectors. Next, to improve the diversity of contrastive pairs and further supplement the data, 
we innovatively use mixup techniques \cite{zhang2018mixup} to simultaneously generate \textit{hard negative} and \textit{hard positive} samples with hard-to-distinguish embeddings, rather than just relying on hard negative mining in SCL. 
With the embeddings of one training batch of original data, the synthetic data constitute a balanced target sample set for SCL learning and create difficult contrastive pairs for model training to achieve better performance. Our key contributions are as follows:

(1) We propose a novel model, namely, SharpReCL, to efficiently handle imbalanced TC tasks and to overcome the existing challenges of SCL-based models.

(2) 
Our model employs simple-sampling and hard-mixup techniques to generate a balanced sample set for SCL to mitigate the class imbalance problem. Moreover, the SCL and classification branches interact and guide each other via prototype vectors for model training. 

(3) We conduct extensive experiments on several benchmark datasets and SharpReCL achieves excellent results on several imbalanced datasets even compared to LLMs, showing its superiority in handling imbalanced data. 
\section{Related Work}
\noindent \textbf{Text Classification.}
As one of the fundamental tasks in web content mining, TC has been applied to many fields, including sentiment analysis, topic labeling, etc. The development of deep learning has brought about a revolution in TC models, where neural network-based models can automatically extract text features compared to traditional methods based on tedious feature engineering. CNN \cite{kim2014convolutional} and RNN \cite{liu2015multi}, as two representative models of neural networks, have achieved good performance in TC tasks. However, the performance of such models is still limited due to the long-term dependency problem. PLMs based on Transformer architectures \cite{vaswani2017attention}, such as BERT \cite{devlin2018bert} and RoBERTa \cite{liu2019roberta}, that use a large-scale corpus for pretraining have shown superior performance in various downstream tasks. Recently, LLMs, such as GPT-3.5 \cite{ouyang2022training} and Llama \cite{touvron2023llama} trained on massive high-quality datasets unify various NLP tasks into the text generation paradigm and achieve outstanding performance in text understanding.

\noindent \textbf{Contrastive Learning.}
In recent years, self-supervised representation learning has made impressive progress. As a representative approach in this field, CL has been successfully used in several domains \cite{chen2020simple,gao2021simcse}. The idea of CL is to align the features between positive pairs while the features of negative pairs are mutually exclusive to achieve effective representation learning. SCL \cite{khosla2020supervised} generalizes unsupervised contrastive approaches to fully-supervised settings to leverage the available label information. Recent works have successfully introduced SCL into TC tasks. For example, DualCL \cite{chen2022dual} adopts a dual-level CL mechanism to construct label-aware text representations by inserting labels directly in top of the original texts. However, SCL performs poorly when directly applied to imbalanced datasets.

\noindent \textbf{Imbalanced Learning.}
Previous methods specific to the class-imbalanced problem are mainly divided into two categories: reweighting \cite{byrd2019effect,zhao2023imbalanced} and resampling \cite{ando2017deep,jiang2023detecting}. The reweighting methods, such as focal loss \cite{lin2017focal} and dice loss \cite{milletari2016v}, assign different weights to different classes, which means that samples from the minority classes are assigned higher weights while those from the majority classes receive relatively lower weights. The resampling approach, such as SMOTE \cite{chawla2002smote}, is mainly implemented by undersampling the majority classes and oversampling the minority classes. Moreover, logit compensation \cite{menon2020long} uses class occurrence frequency as prior knowledge, thus achieving effective classification.
\section{Methods}
In this section, we first define of the problem and the basics of CL. Next, we analyze why CL further aggravates the imbalance problem. Finally, we introduce details of the proposed SharpReCL. 
\begin{figure*}
    \centering
    \includegraphics[width=\linewidth]{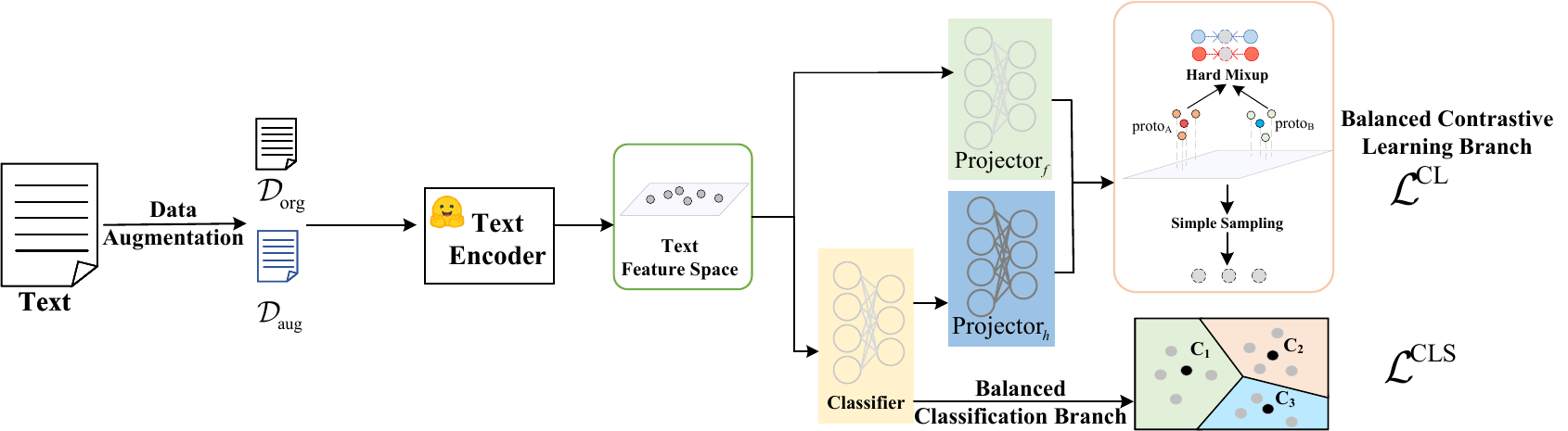}
    \caption{Architecture of SharpReCL (Best viewed in color).}
    \label{flowchart}
\end{figure*}

\subsection{Preliminaries}
\label{preliminary}
\noindent \textbf{Problem Definition}: For multiclass TC tasks with $C$ classes, we assume the given dataset $\{x_{i} ,\ y_{i}\}_{i=1}^{N}$ contains $N$ training examples,  where $x_{i}$ is the text sequence and $1\leqslant y_{i} \leqslant C$ is the label assigned to the input text. Our target is to obtain a function $\phi$ mapping from an input space 
$\mathcal{X}$ to the target space 
$\mathcal{Y}$. Usually, the function $\phi$ is implemented as the composition of an encoder $\Psi$: $\mathcal{X}$ →$\mathcal{Z} \in \mathbb{R}^{h}$ and a linear classifier $\Omega$: $\mathcal{Z}$ → $\mathcal{Y}$. Our model also adopts this structure.

\noindent \textbf{Contrastive Learning}:
Given $N$ training samples  $\{x_{i}\}_{i=1}^{N}$ and their corresponding augmented samples, each sample has at least one augmented sample in the dataset. For an instance $x_{i}$ and its representation $z_{i}$ in a batch $B$, the standard unsupervised CL loss \cite{chen2020simple} can be presented as:
\begin{equation}
    \label{UCL}
    \mathcal{L}^{\text{UCL}}_i =  -\frac{1}{|B|} \log\frac{\exp\left( z_{i} \cdot z_{i}^{+} /\tau \right)}{\sum _{k\in B\backslash \{i\}} \exp( z_{i} \cdot z_{k} /\tau )},
\end{equation}
where $z_{i}^{+}$ denotes the representation of the augmented sample derived from the $i$-th sample $x_{i}$, and $\tau>0$ is the temperature factor that controls tolerance to similar samples.

SCL treats all samples with the same label in the batch as positive and all other samples as negative. Its loss function can be written as:
\begin{equation}
    \label{SCL}
    \begin{aligned}
    \mathcal{L}^{\text{SCL}}_i =  \frac{1}{|B_{y} |-1}\sum _{p\in B_{y} \backslash \{i\}} -\log\frac{\exp( z_{i} \cdot z_{p} /\tau )}{\sum _{k\in B\backslash \{i\}} \exp( z_{i} \cdot z_{k} /\tau )},
    \end{aligned}
\end{equation}
where $B_{y}$ denotes a subset of $B$ that contains all samples of class $y$.

\noindent \textbf{Analysis}:
We suppose the distribution of each class in the original dataset of $N$ samples is $\displaystyle \{\pi _{i}\}, \ 1\leqslant i\leqslant C$. Generally, the imbalance rate in the dataset can be defined as:
\begin{equation}
\label{ir_org}
\text{ir}\ =\frac{\max\{\pi _{i}\} \ }{\min\{\pi _{i}\} \ } \ =\frac{\pi _\text{max}}{\pi _\text{min}}.
\end{equation}

Since SCL works at the sample-pair level, in each batch $B$, we define $P_{i}$ as the number of sample pairs that belong to the same class $i$ and participate in CL. Accordingly, we can define the contrastive imbalance rate as follows:
\begin{equation}
\label{ir_SCL}
    \begin{aligned}
    \text{ir}^{\text{SCL}} \ &=\ \frac{\max\{P_{i}\}}{\min\{P_{i}\}} =\frac{|B|\pi _{\text{max}} \cdot ( |B|\pi _{\text{max}} -1)}{|B|\pi _{\text{min}} \cdot ( |B|\pi _{\text{min}} -1)} \\
    &\approx \frac{( |B|\pi _{\text{max}})^{2}}{( |B|\pi _{\text{min}})^{2}} =\left(\frac{\pi _{\text{max}}}{\pi _{\text{min}}}\right)^{2} ={\text{ir}}^{2}.
    \end{aligned}
\end{equation}
Thus, we find that when SCL is performed, for every batch, the imbalance ratio becomes the square of the original ratio. Since each epoch consists of many batches, assuming that the sampling in each batch is uniform, the imbalance ratio of each epoch also remains $\text{ir}^2$. In summary, since SCL works with sample pairs, such a learning mechanism will significantly amplify the class-imbalanced problem in the original data and create more difficulties for model training.
\subsection{Overview of SharpReCL}
The framework of our model is shown in Fig. \ref{flowchart}. Our model consists of two branches, namely, the balanced classification branch and the balanced CL branch. We first apply conventional data augmentation techniques in NLP (such as back-translation \cite{edunov2018understanding}, random noise injection \cite{xie2019data}, and word substitution \cite{wei2019eda}) to obtain the augmented data $D_{\text{aug}}$ for the original data $D_{\text{org}}$. We default to word substitution, and the results of different augmentation methods can be found in \textbf{Appendix} \ref{data_aug}. 
Here, we denote the texts under two perspectives as $D=D_{\text{org}} \cup D_{\text{aug}}$. Since PLMs such as BERT have shown their superior representation capability, we use PLMs as the 
backbone encoder for text samples in SharpReCL. For each text sequence sample $x_i$, we first obtain its encoded representations using the encoder, \textit{i.e.}, $ \text{feat}_{i}=\text{PLM}(x_{i})\in \mathbb{R}^{h_{1}}$. 
Then, the balanced classification branch performs TC with the derived text representations and obtains class prototype vectors. After that, we project these two types of representations into the same space. 
In this space, the data are expanded by mixup techniques, and then, the balanced CL is performed.
\subsection{Balanced Classification Branch for Class Prototype Generation}
The prototype vector represents each class and acts as a data complement by becoming an instance of the corresponding class in subsequent operations. In SharpReCL, we use the class-specific weights from the backbone classification branch to represent each class, making the class prototype vectors directly learnable. Moreover, these prototype vectors can be corrected by explicitly utilizing labels in the classification task.

Most classification models typically adopt cross-entropy loss, which is suboptimal on imbalanced data since the label distributions introduce bias into the trained model. These models perform well for instances of the majority class but poorly for instances of the minority class. Logit compensation \cite{menon2020long} is an effective method for imbalanced data that overcomes the shortcomings of the previous reweighting methods and increases the classification boundary between classes based on a simple adjustment of the traditional cross-entropy loss with a class prior. Therefore, we adopt logit compensation instead of the conventional cross-entropy loss in our experiments.

Here, we use a linear function to obtain the class logits, where the weights of the linear function are used as class-specific weights. After the procedure of one projection head $\text{proj}_h$, we can obtain the prototype vectors. The above operations can be expressed as:
\begin{equation}
\label{prototype}
    \varphi(\text{feat}) = w^{T}\cdot \text{feat}, \quad
    \text{proto}_{1:C} = \text{proj}_{h}(w),
\end{equation}
where $\varphi(\text{feat}) \in \mathbb{R}^{N\times C}$ denotes the class logits, and  $\text{proto}_{1:C} \in \mathbb{R}^{C\times h_2}$ represents the class prototype vectors. The projection head $\text{proj}_{h}$ is implemented by a two-layer MLP. 
After obtaining class logits, we perform logit compensation for classification, which can be written as:
\begin{equation}
    \label{lc}
    \begin{aligned}
        \mathcal{L}^{\text{CLS}}( y,\ \varphi (\text{feat}))
        =-\log\frac{\exp( \varphi _{y}(\text{feat}) +\delta _{y})}{\sum _{y'\ \in \ [\mathcal{Y}]} \exp( \varphi _{y'}(\text{feat}) +\delta _{y'})}.
    \end{aligned}
\end{equation}
Here, $\delta_y$ stands for the compensation for class $y$, and its value is related to the class-frequency. In SharpReCL, we set the class compensation $\delta_y$ to $\log \text{P}_y$ 
, where $\text{P}_y$ is the class prior of class $y$.
\subsection{Simple-Sampling and Hard-Mixup for the Rebalanced Dataset}
We use the feature projection head $\text{proj}_f$ to map the encoded representations $\text{feat}$ of texts such that the texts' hidden representation $z$ is in the same space as the prototype vector $\text{proto}$. For each text-encoded representation $\text{feat}_i$, this operation can be denoted as:
\begin{equation}
    \label{z}
    z_{i} \ =\ \text{proj}_{f}( \text{feat}_{i}).
\end{equation}

Since prototype vectors are used to represent each class, we address the extreme case where samples from the minority classes do not appear in a batch. This is because we supplement the prototype vectors into each batch of data. We denote the extended dataset as $\hat{D}=\mathrm{Z}\cup \text{proto}$ and then perform $l_2$-normalization on it, where $\mathrm{Z}$ and $\text{proto}$ represent the sets of hidden text vectors and class prototype vectors, respectively. In this section, we further correct the imbalanced dataset by constructing a balanced dataset.

On the one hand, not all samples are helpful for model training, and overly easy samples contribute little to the gradient \cite{robinson2020contrastive}; on the other hand, extreme (particularly hard) samples can also lead to performance degradation \cite{song2022supervised}. Therefore, in SharpReCL, we use a combination of simple-sampling and hard-mixup to construct the balanced dataset.

\noindent \textbf{Simple-Sampling}: As a natural idea, we need to perform resampling on the dataset to expand it. In SCL, positive samples of each class are samples from the same class, and negative samples are other samples from different classes. Therefore, for each class $c,\ 1\leqslant c\leqslant C$, we can construct its positive sample set $ D^{+}$ and negative sample set $ D^{-}$ by simply sampling from $\hat{D}$, which can be written as:
\begin{equation}
\begin{aligned}
    \label{simple-sampling}
    D_{\text{samp},c}^{+} \ =\ \text{Samp}\{z_{i}\} ,\ z_{i} \ \in \hat{D} \ \wedge\ y_{i} =c, \\
    D_{\text{samp},c}^{-} \ =\ \text{Samp}\{z_{i}\} ,\ z_{i} \ \in \hat{D} \ \wedge\ y_{i} \neq c,
\end{aligned}
\end{equation}
where $\text{Samp}\{\cdot\}$ denotes the sampling operation.

\noindent \textbf{Hard-Mixup}: We use the prototype vectors as a criterion to measure the learning difficulty of the samples to obtain the hard sample set. For each class $c,\ 1\leqslant c\leqslant C$ and its prototype vector $\text{proto}_c$, we consider the top-\textit{k} positive samples from class $c$ that are not similar to $\text{proto}_c$ as hard positive samples, and the top-\textit{k} negative samples from other classes that are similar to $\text{proto}_c$ as hard negative samples, which can be denoted as:
\begin{equation}
\begin{aligned}
    \label{hard_set}
    D_{\text{hard},c}^{+} &=\text{top-\textit{k}}\{-\text{proto}_{c} \cdot z_{i},z_{i}\} ,z_{i} \in \hat{D} \ \wedge \ y_{i} =c, \\
    D_{\text{hard},c}^{-} &=\text{top-\textit{k}}\{\text{proto}_{c} \cdot z_{i} ,z_{i}\} ,z_{i} \in \hat{D} \  \wedge \ y_{i} \neq c.
\end{aligned}
\end{equation}

To increase sample diversity and to improve learning efficiency, we need to further amplify the hard samples obtained by previous procedures. 
In SharpReCL, we apply linear interpolation to further augment the data in hard sample sets. This operation is denoted as:
\begin{equation}
\begin{aligned}
    \label{hard-mixup}
    D_{\text{syn},c}^{+} =\left\{\frac{\tilde{z}_{k}^{+}}{||\tilde{z}_{k}^{+} ||}\right\} ,\ \tilde{z}_{k}^{+} =\alpha _{k} z_{i}^{+} +( 1-\alpha _{k}) z_{j}^{+}, \\
\ D_{\text{syn},c}^{-} =\left\{\frac{\tilde{z}_{k}^{-}}{||\tilde{z}_{k}^{-} ||}\right\} ,\ \tilde{z}_{k}^{-} =\beta _{k} z_{i}^{-} +( 1-\beta _{k}) z_{j}^{-},
\end{aligned}
\end{equation}
where ($z_i^+, z_j^+$) and ($z_i^-, z_j^-$) are randomly sampled from the corresponding hard sample sets $D_{\text{hard},c}^{+}$ and $D_{\text{hard},c}^{-}$, and $\alpha _{k}, \beta _{k} \sim \text{Beta}\left( \lambda ,\lambda \right)$ are in the range of $(0,1)$. By default, we set $\lambda$ to 0.5.

In this way, for each class $c$, we construct a balanced positive sample set $\tilde{D}_{c}^{+}\!=\!D_{\text{samp},c}^{+} \!\cup \!D_{\text{syn},c}^{+}$ and a balanced negative sample set $\tilde{D}_{c}^{-}\!=\!D_{\text{samp},c}^{-} \!\cup \!D_{\text{syn}, c}^{-}$. The rebalanced dataset $\tilde{D}$ for all classes can be represented as $\tilde{D}\!=\!\{\tilde{D}_c^{+}\!\cup \tilde{D}_c^{-}\}_c^{C}\!=\!D_{\text{samp}}\!\cup D_{\text{syn}}$.
\subsection{Contrastive Learning Branch with a Rebalanced Dataset}
One issue of concern is the control of the ratio of the hard-mixup dataset $D_{\text{syn}}=\{D_{\text{syn}, c}^{+}\cup D_{\text{syn}, c}^{-}\}_c^C$ to the balanced dataset $\tilde{D}$. Since the feature extraction ability of the model becomes stronger with increasing iterations during the training process, we set the ratio for $D_{\text{syn}}$ to $0.5+\frac{t}{T} \times 0.5$, with $t$ being the current number of iterations and $T$ being the total number of iterations. In this way, the difficulty of the balanced sample set we construct is flexible and controllable, which becomes more challenging during the training process and facilitates the training and learning of the model.

Note that we treat the sample pairs in traditional CL as two parts, $i.e.$, the anchor sample and the target sample. The anchor sample comes from the original samples in the batch together with prototype vectors, $\textit{i.e.}$, $ \hat{D}$, and the target sample set consists of $\hat{D}$ and our constructed sample set $\tilde{D}$. Therefore, the original class imbalance problem still exists in the anchor samples. 

To alleviate this problem, we add the class prior $\delta$ used in Eq. \ref{lc} to the original SCL to correct $\mathcal{L}^{\text{SCL}}$ in Eq. \ref{SCL} based on the idea of reweighting. The concrete objective function can be written as: 
\begin{equation}
\begin{aligned}
    \label{CL}
    &\mathcal{L}^{\text{CL}}_i \!= -\frac{\delta _{y}}{|\hat{D} |}
    \!\sum\nolimits_{p \in \tilde{D}_{y}^{+} \cup  \hat{D}_{y} \!\backslash \{i\}} \log\frac{\exp( z_{i} \cdot z_{p} /\tau )}{\!\sum\nolimits _{k \in \tilde{D}_{y} \cup \hat{D}\!\backslash \{i\}} \exp( z_{i} \cdot z_{k} /\tau )}, \\
    &\mathcal{L}^{\text{CL}} \ =\ \sum\nolimits_{i\in \hat{D} }\mathcal{L}_{i}^{\text{CL}}.
\end{aligned}
\end{equation}
The generation of hard samples can alleviate the issue of vanishing gradients and is beneficial for CL training, as supported by mathematical analysis in \textbf{Section} \ref{math_analysis}. 

Finally, we have the following loss for training:
\begin{equation}
    \label{all_loss}
    \mathcal{L}^{\text{overall}} = \mathcal{L}^{\text{CLS}} \ +\ \mu \mathcal{L}^{\text{CL}},
\end{equation}
where $\mu$ is the hyperparameter that controls the impact of CL branch, and this branch only intends for the backbone to learn the desired feature embeddings. We feed the obtained test text embeddings to the classification branch during the model inference stage to evaluate the model performance. Moreover, both the classification branch and the CL branch influence the learnable prototype vectors, connecting to each other through these vectors and facilitating explicit interaction and mutual guidance. 


The concrete training procedure of our model can be found in Algorithm \ref{pseudo}.
\begin{algorithm}[ht]
\renewcommand{\algorithmicrequire}{\textbf{Input:}}
\renewcommand{\algorithmicensure}{\textbf{Output:}}
\caption{SharpReCL Training}
\label{pseudo}
\begin{algorithmic}[1]
\REQUIRE The text dataset $D=D_{org}\cup D_{aug}=\{x_i,y_i\}_i^{2N}$, the text encoder $PLM(\cdot)$, the linear function $\varphi(\cdot)$, projection heads $proj_f(\cdot)$ and $proj_h(\cdot)$
    \STATE Initialize the network parameters
    \FOR{$t=1,2,\dots, T$ }{
    \STATE Obtain texts' encoded embeddings $feat$ using $PLM(\cdot)$
    \STATE Obtain the class logits and prototype vectors $proto$ with $\varphi(\cdot)$ and $proj_h(\cdot)$ in Eq.\ref{prototype}
    \STATE Obtain hidden embeddings of texts with $proj_f(\cdot)$ in Eq.\ref{z}
    \STATE Compute $|D_{samp}|$ and $|D_{syn}|$ 
    \STATE Perform Simple-Samping to obtain $D_{samp}$ with Eq.\ref{simple-sampling}
    \STATE Perform Hard-Mixup to obtain $D_{hard}$ with Eq.\ref{hard_set}
    \STATE Apply Mixup techniques to obtain $D_{syn}$ with Eq.\ref{hard-mixup}
    \STATE Obtain the re-balanced dataset $\tilde{D}=D_{samp}\cup D_{syn}$
    \STATE Compute the classification loss $\mathcal{L}^{CLS}$ with Eq.\ref{lc}
    \STATE Compute the CL loss $\mathcal{L}^{CL}$
    \STATE Compute the overall loss $\mathcal{L}^{overall}$ with Eq.\ref{all_loss}
    \STATE Optimize the model by backpropagation of $\mathcal{L}^{overall}$  
    }
    \ENDFOR
\ENSURE The well-trained SharpReCL
\end{algorithmic}
\end{algorithm}

\subsection{Mathematical Analysis with Hard Samples}
\label{math_analysis}
In this section, we provide a detailed mathematical analysis of why generating hard samples can alleviate the vanishing gradient problem and benefit CL training. 

For the used supervised contrastive loss as shown in Eq. \ref{CL}, we can obtain the gradient of $\mathcal{L}_i^{CL}$ with respect to node embeddings $z_k$ as follows:
\begin{equation}
\label{gradient}
\begin{aligned}
    &\frac{\partial \mathcal{L}_i^{CL}}{\partial z_i}=\frac{1}{\tau}\bigl[\sum_{p \in \tilde{D}_{y}^{+} \cup  \hat{D}_{y} \backslash \{i\}}z_p(P_{i,p}-\frac{1}{|\hat{D}|}) +\sum_{k \in \tilde{D}_{y}^-  \cup \hat{D} \backslash \{i\cup \hat{D}_{y}\}}z_k P_{i,k}\bigr]\\
   &P_{i,k}=\frac{\exp(z_i\cdot z_k/\tau)}{\sum\limits_{k \in \tilde{D}_{y} \cup \hat{D} \backslash \{i\}}\exp(z_i\cdot z_k/\tau)}
\end{aligned}
\end{equation}

According to Eq. \ref{gradient}, the gradient of $\mathcal{L}_k$ with respect to $z_k$ consists of two parts. The first part is the gradient provided by positive sample pairs, and the second part is the gradient provided by negative sample pairs. When positive pairs are simple, it causes $P_{i,p}\rightarrow\frac{1}{|\hat{D|}}$, whereas when negative pairs are particularly simple, it causes $P_{i,k}\rightarrow 0$. Both scenarios can lead to gradient vanishing, causing unstable model training and damaging the learning process of the entire model. 

However, we only generate hard positive and negative samples, thus providing larger gradients for the contrastive objective. It helps prevent the aforementioned cases and promotes more stable model training.
\section{Experiment}
\label{experiment}
\noindent \textbf{Dataset.}
We employ six widely used real-world datasets to evaluate the effectiveness of our model, where the train/test split of datasets is the same as previous studies \cite{tang2015pte}. They are \textbf{R52} \cite{liu2021deep}, \textbf{Ohsumed} \cite{linmei2019heterogeneous}, \textbf{TREC} \cite{li2002learning}, \textbf{DBLP} \cite{tang2015pte}, \textbf{Biomedical} \cite{xu2017self}, and \textbf{CR} \cite{ding2008holistic}, with detailed descriptions provided in \textbf{Appendix} \ref{des_dataset}. The detailed statistics of these datasets are summrized in Table \ref{data_table}. We use the original R52 and Ohsumed datasets directly, since they are extremely \textbf{imbalanced}. Other datasets are relatively \textbf{balanced}. There is only one category in the TREC with few samples. Thus the calculated ir is large, but we still treat it as a balanced dataset, which needs preprocessing. We create the imbalanced version of balanced datasets by reducing the training samples of each sorted class with a deterministic ir, which is a common practice in imbalanced learning \cite{cui2019class}. For example, if we set ir to 10, the number of each category in a balanced dataset with three categories is 1000, 900, and 890, each category in the imbalanced version is 1000, 100, and 10.

\begin{table}[ht]
\centering
\caption{Detailed statistics of the datasets.}
\label{data_table}
\resizebox{0.45\textwidth}{!}{%
\begin{tabular}{c|cccccc}
\toprule
Dataset    & \#Train & \#Test  & \#Word  & Avg. Len & \#Class & $ir$   \\ \midrule
R52  & 6,532  & 2,568 & 8,892 & 69.8   & 52     & 1307.7 \\
Ohsumed    & 3,353  & 4,043  & 14,157 & 6.8   & 23     & 65.0 \\
TREC       & 5,452  & 500    & 8,751  & 11.3     & 6      & 14.5 \\
DBLP       & 61,422 & 20,000 & 22.27  & 8.5      & 6      & 4.0  \\
Biomedical & 17,976 & 1,998  & 18,888 & 12.9    & 20     & 1.1  \\
CR         & 3,394  & 376    & 5,542  & 18.2     & 2      & 1.8  \\ \bottomrule
\end{tabular}%
}
\end{table}

\noindent \textbf{Baseline.}
We compare the proposed model with the following four types of competitive baselines. \textit{Pretrained language models and their imbalanced learning versions} include \textbf{BERT} \cite{devlin2018bert}, \textbf{BERT+Dice Loss} \cite{li2020dice}, \textbf{BERT+Focal Loss} \cite{lin2017focal}, and \textbf{RoBERTa} \cite{liu2019roberta}. \textit{Contrastive learning models} contain \textbf{SimCSE} \cite{gao2021simcse}, \textbf{SCLCE} \cite{gunel2020supervised}, \textbf{SPCL} \cite{song2022supervised}, and \textbf{DualCL} \cite{chen2022dual}. \textit{Imbalanced contrastive learning models} include \textbf{Hybrid-SC} \cite{wang2021contrastive}, \textbf{BCL} \cite{zhu2022balanced}, and \textbf{TSC} \cite{li2022targeted}. \textit{Large language models} consist of \textbf{GPT-3.5} \cite{ouyang2022training}, \textbf{Bloom-7.1B} \cite{workshop2022bloom}, \textbf{Llama2-7B} \cite{touvron2023llama}, and \textbf{Llama3-8B} \cite{llama3modelcard}. 

\noindent \textbf{Implementation Details.}
In the SharpReCL, we utilize the pre-trained BERT base-uncased model as the text encoder by using AdamW with a learning rate of 5e-5. The weight decay and batch size are set to 5e-4 and 128, respectively. We implement the projection head by an MLP with one hidden layer activated by ReLU. We default to word substitution for augmenting sentences. The temperature $\tau$ of SCL is chosen from \{0.3, 0.5, 1\}. Moreover, the number of hard positive and hard negative samples per class in Eq.\ref{hard_set} is uniformly 20, \textit{i.e.}, $|D_{\text{hard},c}^{+}|=|D_{\text{hard},c}^{-}|=20$. The number of generated rebalanced positive and negative samples when using mixup for each class $c$ are 10 and 500, \textit{i.e.}, $\tilde{D}_c^{+}=10$ and $\tilde{D}_c^{-}=500$. The parameter $\mu$ for controlling the loss function is 1. To accelerate model training, we use two 24GB Nvidia GeForce RTX 3090 GPUs. 

For SimCSE, similar to SCLCE, we use a linear combination of unsupervised CL loss and CE loss to fully utilize the labeled data for model training. For RoBERTa, we adopt the RoBERTa-base version. For imbalanced CL methods, they are originally designed for images. To adapt these methods to the texts, we replace their feature extractors with a text encoder (BERT). For LLMs, we fine-tune GPT-3.5 using the training data from the evaluation dataset through the fine-tuning interface provided by OpenAI. For Bloom-7.1B, Llama2-7B, and Llama3-8B, we perform full-scale fine-tuning using the training data. To reduce GPU memory usage, we employ LoRA \cite{hu2021lora} and 4-bit quantization techniques through the Parameter-Efficient Fine-Tuning (PEFT) method provided by Hugging Face. The prompts used for text classification in Bloom-7.1B, Llama2-7B, and Llama3-8B are the same as those used for GPT-3.5. 
For other remaining baselines, we use the open-source codes and adopt the parameters suggested by their original papers.

The detailed descriptions of baselines are provided in \textbf{Appendix} \ref{des_baseline}. We employ the accuracy (Acc) and macro-f1 score (F1) to examine all the models' performance. All experiments are conducted ten times to obtain average metrics for statistical significance.
\section{Results}

\begin{table*}[ht]
\centering
\caption{The results (\%) of Acc and Macro-F1 scores of various experimental settings. Our model performance is a statistically significant improvement over the best baseline, with a p-value smaller than 0.001. ``Avg. metrics'' is the average model performance across multiple datasets. ``org$^\ast$'' represents the case where the classes are relatively balanced. Underline: runner-up.}
\label{res_one}
\resizebox{\textwidth}{!}{%
\begin{tabular}{@{}cc|c|cccc|cccc|ccc|c@{}}
\toprule
\multicolumn{1}{c|}{Dataset}                     & ir                          & Metric & BERT                                   & \begin{tabular}[c]{@{}c@{}}BERT+\\ Dice Loss\end{tabular} & \begin{tabular}[c]{@{}c@{}}BERT+\\ Focal Loss\end{tabular} & RoBERTa                      & SimCSE                     & SCLCE                                  & SPCL                       & DualCL                                 & Hybrid-SC                  & BCL                                    & TSC                                    & Ours                                   \\ \midrule
\multicolumn{1}{c|}{\multirow{2}{*}{R52}}        & \multirow{2}{*}{org}        & Acc    & 95.55$\pm$\scriptsize 0.36             & 95.48$\pm$\scriptsize 0.29                                & \underline{95.66}$\pm$\scriptsize 0.25                                 & 95.56$\pm$\scriptsize 0.22   & 89.35$\pm$\scriptsize 0.20 & 95.51$\pm$\scriptsize 0.35             & 94.29$\pm$\scriptsize 0.36 & 94.16$\pm$\scriptsize 0.16             & 95.49$\pm$\scriptsize 0.51 & 95.56$\pm$\scriptsize 0.36             & 95.64$\pm$\scriptsize 0.32             & \textbf{96.11}$\pm$\scriptsize 0.36    \\
\multicolumn{1}{c|}{}                            &                             & F1     & 82.88$\pm$\scriptsize 0.28             & 81.58$\pm$\scriptsize 0.32                                & 83.19$\pm$\scriptsize 0.26                                 & 82.33$\pm$\scriptsize 0.35   & 79.22$\pm$\scriptsize 0.15 & 83.64$\pm$\scriptsize 0.39             & 83.19$\pm$\scriptsize 0.32 & 73.92$\pm$\scriptsize 0.12             & 83.26$\pm$\scriptsize 0.50 & \underline{83.59}$\pm$\scriptsize 0.29             & 82.90$\pm$\scriptsize 0.55             & \textbf{84.83}$\pm$\scriptsize 0.29    \\ \midrule
\multicolumn{1}{c|}{\multirow{2}{*}{Ohsumed}}    & \multirow{2}{*}{org}        & Acc    & 65.71$\pm$\scriptsize 0.30             & 65.94$\pm$\scriptsize 0.20                                & 65.22$\pm$\scriptsize 0.26                                 & 66.06$\pm$\scriptsize 0.23   & 49.30$\pm$\scriptsize 0.32 & 66.59$\pm$\scriptsize 0.35             & 67.20$\pm$\scriptsize 0.42 & 26.44$\pm$\scriptsize 0.35             & 66.32$\pm$\scriptsize 0.42 & 66.39$\pm$\scriptsize 0.62             & \underline{66.66}$\pm$\scriptsize 0.72 & \textbf{68.44}$\pm$\scriptsize 0.22    \\
\multicolumn{1}{c|}{}                            &                             & F1     & \underline{57.36}$\pm$\scriptsize 0.31 & 55.59 $\pm$\scriptsize 0.36                               & 55.00$\pm$\scriptsize 0.28                                 & 56.95$\pm$\scriptsize 0.26   & 36.00$\pm$\scriptsize 0.40 & 56.25$\pm$\scriptsize 0.39             & 56.42$\pm$\scriptsize 0.42 & 14.75$\pm$\scriptsize 0.55             & 58.79$\pm$\scriptsize 0.66 & 58.85$\pm$\scriptsize 0.72             & 57.92$\pm$\scriptsize 0.69             & \textbf{60.82}$\pm$\scriptsize 0.20    \\ \midrule
\multicolumn{1}{c|}{\multirow{8}{*}{TREC}}       & \multirow{2}{*}{org$^\ast$} & Acc    & 97.60$\pm$\scriptsize 0.39             & 97.60$\pm$\scriptsize 0.42                                & 97.20$\pm$\scriptsize 0.56                                 & \underline{97.61}$\pm$\scriptsize 0.55   & 88.80$\pm$\scriptsize 0.62 & 97.60$\pm$\scriptsize 0.38             & 96.50$\pm$\scriptsize 0.95 & \textbf{97.80}$\pm$\scriptsize 1.25    & 94.21$\pm$\scriptsize 1.59 & 95.60$\pm$\scriptsize 1.65             & 96.62$\pm$\scriptsize 1.62             & 97.20$\pm$\scriptsize 0.95             \\
\multicolumn{1}{c|}{}                            &                             & F1     & 96.99$\pm$\scriptsize 0.52             & 96.97$\pm$\scriptsize 0.55                                & 96.13$\pm$\scriptsize 0.65                                 & 97.00$\pm$\scriptsize 0.62   & 85.90$\pm$\scriptsize 0.72 & 97.01$\pm$\scriptsize 0.42             & 95.50$\pm$\scriptsize 1.20 & \underline{97.38}$\pm$\scriptsize 1.72 & 96.92$\pm$\scriptsize 1.86 & 96.36$\pm$\scriptsize 1.96             & 97.09$\pm$\scriptsize 1.92             & \textbf{97.63}$\pm$\scriptsize 0.82    \\ \cmidrule(l){2-15} 
\multicolumn{1}{c|}{}                            & \multirow{2}{*}{50}         & Acc    & 93.40$\pm$\scriptsize 0.66             & 93.60$\pm$\scriptsize 0.50                                & 94.40$\pm$\scriptsize 0.52                                 & 92.60$\pm$\scriptsize 0.39   & 74.00$\pm$\scriptsize 0.29 & 94.20$\pm$\scriptsize 0.30             & 93.52$\pm$\scriptsize 1.25 & 92.80$\pm$\scriptsize 1.55             & 93.62$\pm$\scriptsize 1.99 & 93.40$\pm$\scriptsize 1.76             & 94.02$\pm$\scriptsize 2.25             & \textbf{94.80}$\pm$\scriptsize 0.55 \\
\multicolumn{1}{c|}{}                            &                             & F1     & 93.30$\pm$\scriptsize 0.65           & 92.36$\pm$\scriptsize 0.69                              & \underline{94.26}$\pm$\scriptsize 0.40                   & 92.50$\pm$\scriptsize 0.41 & 76.00$\pm$\scriptsize 0.32 & 93.97$\pm$\scriptsize 0.39             & 92.40$\pm$\scriptsize 1.52 & 91.05$\pm$\scriptsize 1.62             & 93.79$\pm$\scriptsize 2.01 & 92.00$\pm$\scriptsize 2.12             & 93.46$\pm$\scriptsize 2.29             & \textbf{94.83}$\pm$\scriptsize 0.52    \\ \cmidrule(l){2-15} 
\multicolumn{1}{c|}{}                            & \multirow{2}{*}{20}         & Acc    & 96.60$\pm$\scriptsize 0.49            & 96.60$\pm$\scriptsize 0.28                                & 95.80$\pm$\scriptsize 0.46                                 & 96.60$\pm$\scriptsize 0.53   & 82.20$\pm$\scriptsize 0.35 & 96.80$\pm$\scriptsize 0.60             & 95.52$\pm$\scriptsize 0.62 & 96.40$\pm$\scriptsize 1.79             & 95.79$\pm$\scriptsize 2.15 & 95.90$\pm$\scriptsize 2.39             & 96.60$\pm$\scriptsize 2.06             & \textbf{97.00}$\pm$\scriptsize 0.52    \\
\multicolumn{1}{c|}{}                            &                             & F1     & 96.05$\pm$\scriptsize 0.56             & 96.25$\pm$\scriptsize 0.19                                & 95.53$\pm$\scriptsize 0.55                                 & 95.31$\pm$\scriptsize 0.52   & 82.00$\pm$\scriptsize 0.38 & \underline{96.32}$\pm$\scriptsize 0.71 & 94.19$\pm$\scriptsize 0.63 & 95.25$\pm$\scriptsize 1.59             & 95.86$\pm$\scriptsize 1.90 & 95.25$\pm$\scriptsize 2.25             & 96.17$\pm$\scriptsize 2.42             & \textbf{96.45}$\pm$\scriptsize 0.21    \\ \cmidrule(l){2-15} 
\multicolumn{1}{c|}{}                            & \multirow{2}{*}{10}         & Acc    & 96.60$\pm$\scriptsize 0.42             & 97.00$\pm$\scriptsize 0.30                                & 96.80$\pm$\scriptsize 0.27                                 & 97.00$\pm$\scriptsize 0.24   & 85.00$\pm$\scriptsize 0.25 & 97.00$\pm$\scriptsize 0.72             & 96.70$\pm$\scriptsize 0.75 & \textbf{97.60}$\pm$\scriptsize 1.66    & 96.12$\pm$\scriptsize 1.22 & 95.40$\pm$\scriptsize 2.12             & 96.80$\pm$\scriptsize 1.62             & 97.20$\pm$\scriptsize 0.35             \\
\multicolumn{1}{c|}{}                            &                             & F1     & 95.38$\pm$\scriptsize 0.55             & 95.99$\pm$\scriptsize 0.37                                & 96.41$\pm$\scriptsize 0.29                                 & 96.55$\pm$\scriptsize 0.46   & 82.20$\pm$\scriptsize 0.70 & 95.64$\pm$\scriptsize 0.82             & 95.72$\pm$\scriptsize 0.71 & \underline{97.34}$\pm$\scriptsize 1.72 & 96.09$\pm$\scriptsize 1.51 & 95.89$\pm$\scriptsize 2.22             & 96.32$\pm$\scriptsize 1.55             & \textbf{97.53}$\pm$\scriptsize 0.39    \\ \midrule
\multicolumn{1}{c|}{\multirow{8}{*}{DBLP}}       & \multirow{2}{*}{org$^\ast$} & Acc    & 79.62$\pm$\scriptsize 0.66            & 80.09$\pm$\scriptsize 0.38                               & 79.73$\pm$\scriptsize 0.53                                & 79.85$\pm$\scriptsize 0.32  & 68.50$\pm$\scriptsize 0.39 & 79.96$\pm$\scriptsize 0.59             & 79.85$\pm$\scriptsize 0.66 & \textbf{80.52}$\pm$\scriptsize 1.73             & 77.96$\pm$\scriptsize 1.55 & 78.76$\pm$\scriptsize 2.05             & 78.20$\pm$\scriptsize 0.85             & \underline{80.15}$\pm$\scriptsize 0.39             \\
\multicolumn{1}{c|}{}                            &                             & F1     & 76.47$\pm$\scriptsize 0.18             & 77.27$\pm$\scriptsize 0.15                                & 76.57$\pm$\scriptsize 0.25                                 & 76.99$\pm$\scriptsize 0.45   & 64.40$\pm$\scriptsize 0.23 & 76.84$\pm$\scriptsize 0.60             & 77.20$\pm$\scriptsize 0.68 & \textbf{77.67}$\pm$\scriptsize 1.22             & 77.16$\pm$\scriptsize 1.29 & 76.08$\pm$\scriptsize 2.39             & 76.19$\pm$\scriptsize 1.25             & \underline{77.59}$\pm$\scriptsize 0.28             \\ \cmidrule(l){2-15} 
\multicolumn{1}{c|}{}                            & \multirow{2}{*}{50}         & Acc    & 74.53$\pm$\scriptsize 0.38             & 74.34$\pm$\scriptsize 0.39                                & 73.94$\pm$\scriptsize 0.42                                 & 74.19$\pm$\scriptsize 0.53   & 62.40$\pm$\scriptsize 0.30 & 74.49$\pm$\scriptsize 0.90             & 74.96$\pm$\scriptsize 0.78 & \underline{75.72}$\pm$\scriptsize 1.85             & 75.19$\pm$\scriptsize 1.82 & 75.60$\pm$\scriptsize 2.19             & 75.50$\pm$\scriptsize 2.22             & \textbf{76.30}$\pm$\scriptsize 0.25             \\
\multicolumn{1}{c|}{}                            &                             & F1     & 68.60$\pm$\scriptsize 0.32             & 67.20$\pm$\scriptsize 0.45                                & 67.99$\pm$\scriptsize 0.39                                 & 67.03$\pm$\scriptsize 0.59   & 51.30$\pm$\scriptsize 0.33 & 67.75$\pm$\scriptsize 0.92             & 67.70$\pm$\scriptsize 0.87 & \underline{69.57}$\pm$\scriptsize 1.99             & 69.22$\pm$\scriptsize 2.01 & 69.10$\pm$\scriptsize 2.32             & 69.30$\pm$\scriptsize 2.55             & \textbf{71.15}$\pm$\scriptsize 0.36             \\ \cmidrule(l){2-15} 
\multicolumn{1}{c|}{}                            & \multirow{2}{*}{20}         & Acc    & 77.20$\pm$\scriptsize 0.93             & 76.95$\pm$\scriptsize 0.65                                & 77.11$\pm$\scriptsize 0.16                                 & 76.70$\pm$\scriptsize 0.29   & 65.10$\pm$\scriptsize 0.35 & 77.65$\pm$\scriptsize 0.26             & 77.72$\pm$\scriptsize 0.73 & \underline{78.20}$\pm$\scriptsize 2.01             & 76.12$\pm$\scriptsize 2.05 & 76.40$\pm$\scriptsize 1.88             & 76.62$\pm$\scriptsize 2.16             & \textbf{78.72}$\pm$\scriptsize 0.31             \\
\multicolumn{1}{c|}{}                            &                             & F1     & 72.93$\pm$\scriptsize 1.11             & 72.69$\pm$\scriptsize 0.68                                & 72.71$\pm$\scriptsize 0.23                                 & 72.15$\pm$\scriptsize 0.35   & 57.00$\pm$\scriptsize 0.39 & 73.16$\pm$\scriptsize 0.30             & 73.12$\pm$\scriptsize 0.76 & \underline{74.07}$\pm$\scriptsize 1.98             & 72.16$\pm$\scriptsize 2.15 & 73.10$\pm$\scriptsize 1.93             & 73.25$\pm$\scriptsize 2.55             & \textbf{75.13}$\pm$\scriptsize 0.38             \\ \cmidrule(l){2-15} 
\multicolumn{1}{c|}{}                            & \multirow{2}{*}{10}         & Acc    & 78.86$\pm$\scriptsize 0.99             & 78.35$\pm$\scriptsize 0.82                                & 78.70$\pm$\scriptsize 0.36                                 & 78.21$\pm$\scriptsize 0.39   & 67.10$\pm$\scriptsize 0.38 & 78.96$\pm$\scriptsize 0.86             & 78.42$\pm$\scriptsize 0.77 & \underline{79.03}$\pm$\scriptsize 2.66             & 77.52$\pm$\scriptsize 2.18 & 77.92$\pm$\scriptsize 2.08 & 78.22$\pm$\scriptsize 2.60             & \textbf{79.59}$\pm$\scriptsize 0.41             \\
\multicolumn{1}{c|}{}                            &                             & F1     & 75.16$\pm$\scriptsize 1.22             & 74.78$\pm$\scriptsize 0.89                                & 75.30$\pm$\scriptsize 0.32                                 & 74.52$\pm$\scriptsize 0.55   & 61.00$\pm$\scriptsize 0.42 & \underline{75.92}$\pm$\scriptsize 0.88             & 75.42$\pm$\scriptsize 0.79 & 75.81$\pm$\scriptsize 2.72             & 73.52$\pm$\scriptsize 2.02 & 73.42$\pm$\scriptsize 2.11 & 73.06$\pm$\scriptsize 2.69             & \textbf{76.75}$\pm$\scriptsize 0.46             \\ \midrule
\multicolumn{1}{c|}{\multirow{8}{*}{Biomedical}} & \multirow{2}{*}{org$^\ast$} & Acc    & 71.22$\pm$\scriptsize 0.82             & 70.76$\pm$\scriptsize 0.73                                & 70.59$\pm$\scriptsize 0.30                                 & 70.78$\pm$\scriptsize 0.62   & 62.80$\pm$\scriptsize 0.29 & \underline{71.42}$\pm$\scriptsize 0.60             & 70.92$\pm$\scriptsize 0.66 & 70.90$\pm$\scriptsize 1.89             & 69.51$\pm$\scriptsize 2.29 & 69.90$\pm$\scriptsize 2.09 & 69.39$\pm$\scriptsize 2.51             & \textbf{72.52}$\pm$\scriptsize 0.39             \\
\multicolumn{1}{c|}{}                            &                             & F1     & 71.10$\pm$\scriptsize 0.86             & 70.69$\pm$ \scriptsize 0.75                              & 70.32$\pm$\scriptsize 0.39                                 & 70.56$\pm$\scriptsize 0.65   & 61.85$\pm$\scriptsize 0.32 & \underline{71.32}$\pm$\scriptsize 0.75             & 69.39$\pm$\scriptsize 0.72 & 70.81$\pm$\scriptsize 1.92             & 69.66$\pm$\scriptsize 2.19 & 70.02$\pm$\scriptsize 2.19 & 70.50$\pm$\scriptsize 2.59             & \textbf{72.65}$\pm$\scriptsize 0.26             \\ \cmidrule(l){2-15} 
\multicolumn{1}{c|}{}                            & \multirow{2}{*}{50}         & Acc    & 59.17$\pm$\scriptsize 0.56             & 59.32$\pm$\scriptsize 0.79                                & 55.52$\pm$\scriptsize 0.19                                 & 58.14$\pm$\scriptsize 0.51   & 58.22$\pm$\scriptsize 0.33 & 59.28$\pm$\scriptsize 1.01             & \underline{60.16}$\pm$\scriptsize 0.69 & 59.42$\pm$\scriptsize 1.90 & 57.79$\pm$\scriptsize 2.35 & 58.10$\pm$\scriptsize 2.30             & 58.92$\pm$\scriptsize 2.22             & \textbf{61.56}$\pm$\scriptsize 0.28    \\
\multicolumn{1}{c|}{}                            &                             & F1     & 57.72$\pm$\scriptsize 0.55             & 58.89$\pm$\scriptsize 0.82                    & 53.76$\pm$\scriptsize 0.20                                 & 57.44$\pm$\scriptsize 0.55   & 56.45$\pm$\scriptsize 0.36 & 58.45$\pm$\scriptsize 1.09             & \underline{59.05}$\pm$\scriptsize 0.89 & 58.72$\pm$\scriptsize 1.99             & 57.62$\pm$\scriptsize 2.16 & 58.10$\pm$\scriptsize 2.55             & 57.85$\pm$\scriptsize 2.02             & \textbf{60.78}$\pm$\scriptsize 0.12    \\ \cmidrule(l){2-15} 
\multicolumn{1}{c|}{}                            & \multirow{2}{*}{20}         & Acc    & 64.42$\pm$\scriptsize 0.35             & \underline{64.48}$\pm$\scriptsize 0.63                    & 62.67$\pm$\scriptsize 0.29                                 & 62.82$\pm$\scriptsize 0.66   & 54.50$\pm$\scriptsize 0.26 & 64.37$\pm$\scriptsize 0.15             & 64.19$\pm$\scriptsize 0.39 & 63.65$\pm$\scriptsize 1.93             & 64.26$\pm$\scriptsize 2.15 & 64.32$\pm$\scriptsize 1.86             & 63.55$\pm$\scriptsize 2.11             & \textbf{65.47}$\pm$\scriptsize 0.22    \\
\multicolumn{1}{c|}{}                            &                             & F1     & \underline{64.09}$\pm$\scriptsize 0.31 & 64.06$\pm$\scriptsize 0.56                                & 61.92$\pm$\scriptsize 0.22                                 & 62.24$\pm$\scriptsize 0.72   & 54.30$\pm$\scriptsize 0.30 & 63.96$\pm$\scriptsize 0.21             & 63.49$\pm$\scriptsize 0.28 & 63.06$\pm$\scriptsize 1.99             & 63.25$\pm$\scriptsize 2.01 & 62.16$\pm$\scriptsize 2.05             & 62.96$\pm$\scriptsize 2.19             & \textbf{64.83}$\pm$\scriptsize 0.26    \\ \cmidrule(l){2-15} 
\multicolumn{1}{c|}{}                            & \multirow{2}{*}{10}         & Acc    & \underline{66.90}$\pm$\scriptsize 0.39 & 66.74$\pm$\scriptsize 0.70                                & 66.39$\pm$\scriptsize 0.35                                 & 66.73$\pm$\scriptsize 0.82   & 58.20$\pm$\scriptsize 0.30 & 66.81$\pm$\scriptsize 0.66             & 66.10$\pm$\scriptsize 0.58 & 66.45$\pm$\scriptsize 0.86             & 62.26$\pm$\scriptsize 2.19 & 65.80$\pm$\scriptsize 2.09             & 66.62$\pm$\scriptsize 1.99             & \textbf{68.11}$\pm$\scriptsize 0.60    \\
\multicolumn{1}{c|}{}                            &                             & F1     & \underline{66.97}$\pm$\scriptsize 0.31 & 66.69$\pm$\scriptsize 0.69                                & 66.32$\pm$\scriptsize 0.36                                 & 66.52$\pm$\scriptsize 0.89   & 58.30$\pm$\scriptsize 0.39 & 66.77$\pm$\scriptsize 0.69             & 64.92$\pm$\scriptsize 0.66 & 66.50$\pm$\scriptsize 0.90             & 61.19$\pm$\scriptsize 1.98 & 65.02$\pm$\scriptsize 2.26             & 66.22$\pm$\scriptsize 2.05             & \textbf{68.06}$\pm$\scriptsize 0.55    \\ \midrule
\multicolumn{1}{c|}{\multirow{8}{*}{CR}}         & \multirow{2}{*}{org$^\ast$} & Acc    & 92.02$\pm$\scriptsize 0.56             & 92.26$\pm$\scriptsize 0.29                                & 91.02$\pm$\scriptsize 0.39                                 & 92.55$\pm$\scriptsize 0.72   & 86.05$\pm$\scriptsize 0.26 & 91.22$\pm$\scriptsize 0.16             & 91.55$\pm$\scriptsize 0.26 & \underline{92.65}$\pm$\scriptsize 0.77             & \underline{92.16}$\pm$\scriptsize 2.02 & 92.55$\pm$\scriptsize 2.11             & 92.41$\pm$\scriptsize 2.09 & \textbf{93.09}$\pm$\scriptsize 0.30             \\
\multicolumn{1}{c|}{}                            &                             & F1     & 91.39$\pm$\scriptsize 0.59             & 91.60$\pm$\scriptsize 0.19                                & 91.30$\pm$\scriptsize 0.36                                 & 92.03$\pm$\scriptsize 0.77   & 85.92$\pm$\scriptsize 0.32 & 90.62$\pm$\scriptsize 0.22             & 91.36$\pm$\scriptsize 0.28 & 92.16$\pm$\scriptsize 0.79             & 90.02$\pm$\scriptsize 2.19 & 91.01$\pm$\scriptsize 2.19             & 91.03$\pm$\scriptsize 2.16 & \textbf{92.51}$\pm$\scriptsize 0.22             \\ \cmidrule(l){2-15} 
\multicolumn{1}{c|}{}                            & \multirow{2}{*}{50}         & Acc    & 77.39$\pm$\scriptsize 1.01             & \underline{80.59}$\pm$\scriptsize 0.26                                & 77.93$\pm$\scriptsize 0.55                                 & 78.99$\pm$\scriptsize 0.56   & 77.90$\pm$\scriptsize 0.31 & 79.69$\pm$\scriptsize 0.18             & 80.25$\pm$\scriptsize 0.25 & 77.13$\pm$\scriptsize 0.70             & 78.55$\pm$\scriptsize 2.05 & 78.71$\pm$\scriptsize 2.19             & 78.92$\pm$\scriptsize 1.95             & \textbf{84.04}$\pm$\scriptsize 0.25    \\
\multicolumn{1}{c|}{}                            &                             & F1     & 70.38$\pm$\scriptsize 1.06             & 75.37$\pm$\scriptsize 0.32                                & 70.67$\pm$\scriptsize 0.56                                 & 73.83$\pm$\scriptsize 0.59   & 71.10$\pm$\scriptsize 0.35 & 74.46$\pm$\scriptsize 0.22             & 75.62$\pm$\scriptsize 0.30 & 69.50$\pm$\scriptsize 0.72             & 75.29$\pm$\scriptsize 2.02 & 75.46$\pm$\scriptsize 2.25             & \underline{76.41}$\pm$\scriptsize 2.15 & \textbf{81.12}$\pm$\scriptsize 0.26    \\ \cmidrule(l){2-15} 
\multicolumn{1}{c|}{}                            & \multirow{2}{*}{20}         & Acc    & 84.04$\pm$\scriptsize 0.39             & 83.24$\pm$\scriptsize 0.55                                & 84.04$\pm$\scriptsize 0.62                                 & 80.59$\pm$\scriptsize 0.22   & 81.90$\pm$\scriptsize 0.31 & 85.69$\pm$\scriptsize 0.35             & 85.42$\pm$\scriptsize 0.55 & \underline{88.83}$\pm$\scriptsize 0.60             & 85.55$\pm$\scriptsize 1.79 & 84.76$\pm$\scriptsize 1.99 & 85.96$\pm$\scriptsize 1.95    & \textbf{89.36}$\pm$\scriptsize 0.50             \\
\multicolumn{1}{c|}{}                            &                             & F1     & 80.51$\pm$\scriptsize 0.36             & 79.48$\pm$\scriptsize 0.56                                & 80.83$\pm$\scriptsize 0.65                                & 75.37$\pm$\scriptsize 0.23   & 77.40$\pm$\scriptsize 0.36 & 83.35$\pm$\scriptsize 0.36             & 82.56$\pm$\scriptsize 0.59 & \underline{87.20}$\pm$\scriptsize 0.66             & 82.22$\pm$\scriptsize 1.85 & 83.60$\pm$\scriptsize 2.02 & 82.73$\pm$\scriptsize 2.15    & \textbf{88.28}$\pm$\scriptsize 0.52             \\ \cmidrule(l){2-15} 
\multicolumn{1}{c|}{}                            & \multirow{2}{*}{10}         & Acc    & 86.44$\pm$\scriptsize 0.22             & 88.30$\pm$\scriptsize 0.26                                & \underline{90.69}$\pm$\scriptsize 0.60                                 & 90.16$\pm$\scriptsize 0.66   & 83.80$\pm$\scriptsize 0.28 & 88.28$\pm$\scriptsize 0.31             & 89.15$\pm$\scriptsize 0.52 & 90.43$\pm$\scriptsize 0.70             & 89.09$\pm$\scriptsize 2.10 & 89.29$\pm$\scriptsize 1.75 & 89.02$\pm$\scriptsize 2.16             & \textbf{90.96}$\pm$\scriptsize 0.16             \\
\multicolumn{1}{c|}{}                            &                             & F1     & 84.29$\pm$\scriptsize 0.25             & 86.60$\pm$\scriptsize 0.30                                & \underline{89.53}$\pm$\scriptsize 0.55                                 & 88.94$\pm$\scriptsize 0.69   & 80.70$\pm$\scriptsize 0.32 & 87.29$\pm$\scriptsize 0.39             & 88.25$\pm$\scriptsize 0.55 & 89.38$\pm$\scriptsize 0.79             & 87.56$\pm$\scriptsize 2.19 & 87.43$\pm$\scriptsize 1.89 & 87.08$\pm$\scriptsize 2.11             & \textbf{90.21}$\pm$\scriptsize 0.26             \\ \midrule
\multicolumn{2}{c|}{\multirow{2}{*}{Avg. metrics}}                             & Acc    & 80.96                                  & 81.20                                                     & 80.75                                                      & 80.84                        & 71.95                      & 81.42                                  & 81.25                      & 79.34                                  & 80.41                      & 80.79                                  & \underline{81.09}                        & \textbf{82.81}                         \\
\multicolumn{2}{c|}{}                                                          & F1     & 77.87                                  & 78.00                                                     & 77.68                                                      & 77.68                        & 67.84                      & 78.48                                  & 78.08                      & 75.79                                  & 77.97                      & 78.14                                  & \underline{78.38}                        & \textbf{80.62}                         \\ \bottomrule
\end{tabular}%
}
\end{table*}

\begin{table*}[ht]
\centering
\caption{The results of our model and LLMs across different datasets.}
\label{res_llm}
\resizebox{\textwidth}{!}{%
\begin{tabular}{@{}c|cc|cc|cc|cc|cc|cc|cc@{}}
\toprule
Dataset    & \multicolumn{2}{c|}{R52}                                                        & \multicolumn{2}{c|}{Ohsumed}                                                 & \multicolumn{2}{c|}{TREC}                                                    & \multicolumn{2}{c|}{DBLP}                                                       & \multicolumn{2}{c|}{Biomedical}                                           & \multicolumn{2}{c|}{CR}                                                         & \multicolumn{2}{c}{Avg. metrics}  \\ \midrule
Metric     & Acc                                    & F1                                     & Acc                                    & F1                                  & Acc                                    & F1                                  & Acc                                    & F1                                     & Acc                                 & F1                                  & Acc                                    & F1                                     & Acc             & F1              \\ \midrule
GPT-3.5    & 95.27$\pm$\scriptsize 0.55             & 79.15$\pm$\scriptsize 0.52             & 51.84$\pm$\scriptsize 0.45             & 43.83$\pm$\scriptsize 0.65          & 96.62$\pm$\scriptsize 1.25             & 96.26$\pm$\scriptsize 1.52          & 78.52$\pm$\scriptsize 2.39             & 76.52$\pm$\scriptsize 2.42             & 63.26$\pm$\scriptsize 2.11          & 63.19$\pm$\scriptsize 1.98          & 91.09$\pm$\scriptsize 2.10             & 90.56$\pm$\scriptsize 2.19             & 80.35           & 77.65           \\
Bloom-7.1B & 95.42$\pm$\scriptsize 0.36             & 81.29$\pm$\scriptsize 0.29             & 67.54$\pm$\scriptsize 0.62             & 53.52$\pm$\scriptsize 0.72          & \underline{97.40}$\pm$\scriptsize 2.12 & 96.89$\pm$\scriptsize 2.22          & \underline{80.92}$\pm$\scriptsize 2.08 & \underline{77.42}$\pm$\scriptsize 2.11 & 65.80$\pm$\scriptsize 2.09          & 65.02$\pm$\scriptsize 2.26          & \underline{92.29}$\pm$\scriptsize 1.75 & \underline{91.43}$\pm$\scriptsize 1.89 & 81.29           & 77.32           \\
Llama2-7B        & 95.64$\pm$\scriptsize 0.32             & 82.90$\pm$\scriptsize 0.55             & \underline{67.66}$\pm$\scriptsize 0.72 & 55.92$\pm$\scriptsize 0.69          & 96.80$\pm$\scriptsize 1.62             & 96.32$\pm$\scriptsize 1.55          & 80.22$\pm$\scriptsize 2.60             & 77.06$\pm$\scriptsize 2.69             & 66.62$\pm$\scriptsize 1.99          & 66.22$\pm$\scriptsize 2.05          & 92.02$\pm$\scriptsize 2.16             & 91.08$\pm$\scriptsize 2.11             & \underline{82.00} & \underline{78.56} \\
Llama3-8B  & \underline{95.79}$\pm$\scriptsize 0.22 & \underline{83.72}$\pm$\scriptsize 0.35 & 68.02$\pm$\scriptsize 0.26             & 58.59$\pm$\scriptsize 0.19          & 95.20$\pm$\scriptsize 1.66             & 93.59$\pm$\scriptsize 1.35          & \textbf{81.26}$\pm$\scriptsize 2.60    & \textbf{77.75}$\pm$\scriptsize 0.46    & 62.32$\pm$\scriptsize 1.90          & 62.82$\pm$\scriptsize 0.55          & \textbf{93.35}$\pm$\scriptsize 0.66    & \textbf{92.66}$\pm$\scriptsize 0.59    & 81.19           & 78.42           \\ \midrule
Ours       & \textbf{96.11}$\pm$\scriptsize 0.36    & \textbf{84.83}$\pm$\scriptsize 0.29    & \textbf{68.44}$\pm$\scriptsize 0.22    & \textbf{60.82}$\pm$\scriptsize 0.20 & 97.20$\pm$\scriptsize 0.35             & \textbf{97.53}$\pm$\scriptsize 0.39 & 79.59$\pm$\scriptsize 0.41             & 76.75$\pm$\scriptsize 0.46             & \textbf{68.11}$\pm$\scriptsize 0.60 & \textbf{68.06}$\pm$\scriptsize 0.55 & \textbf{90.96}$\pm$\scriptsize 0.16             & \textbf{90.21}$\pm$\scriptsize 0.26             & \textbf{82.81}  & \textbf{80.62}  \\ \bottomrule
\end{tabular}%
}
\end{table*}

\noindent \textbf{Model Performance.} 
We perform extensive experiments on our model and other competitive baselines with six public datasets under different imbalanced settings (R52 and Ohsumed), including original data, ir=50, ir=20 and ir=10, as shown in Table \ref{res_one}. Furthermore, we present the results of our model and other LLMs under ir=50 in Table \ref{res_llm}. Based on the quantitative results, we gain in-depth insights and analyses. First, our model significantly outperforms other baselines in most imbalanced experimental settings if excluding LLMs, illustrating its superiority in modeling imbalanced text data. An important reason is that SharpReCL introduces class prototypes during training, which ensures that all classes appear in every batch, effectively resolving the problem of missing minority classes due to data sampling. Meanwhile, we introduce hard positive and hard negative samples for each class in SCL, which can provide more adequate gradients and thus facilitate model optimization. Based on this, we generate a balanced contrastive queue for each class by using the mixup technique, which effectively alleviates the previous issue that the majority categories dominate the training process in an imbalanced dataset. Additionally, we enable the two learning branches to explicitly interact and mutually guide each other through the prototype vectors, thereby enhancing model training. 

We find other models achieve inferior performance due to serious bias in representation learning. Notably, our model achieves the best performance on several datasets (such as Ohsumed and Biomedical) even when compared to LLMs. One plausible reason is that LLMs lack sufficient domain-specific training data, leading to their inadequate understanding of these texts. We find that SharpReCL does not perform the best on the DBLP compared to LLMs. This can be explained by the fact that, as shown in Table \ref{data_table}, DBLP is significantly larger than other datasets. Consequently, each class in DBLP has enough samples for LLMs to extract meaningful features for downstream classification. 
Although imbalanced contrastive learning models, such as Hybrid-SC and TSC, demonstrate superior performance compared to other categories of models, they still fall behind our model. This performance gap is primarily due to their failure to explicitly exploit hard positive and hard negative samples. Moreover, the imbalanced learning models (\textit{e.g.}, BERT+Dice Loss and BERT+ Focal Loss) and contrastive learning models (\textit{e.g.}, SCLCE, SPCL and DualCL) have comparable performance across all the datasets; the former is dedicated to imbalanced learning, while the latter enjoy the benefits of CL.

\begin{table*}[ht]
\centering
\caption{The results of ablation study.}
\label{ablation}
\small
\begin{tabular}{c|cc|cc|cc|cc|cc|cc}
\toprule
\multirow{2}{*}{Model} &
  \multicolumn{2}{c|}{R52} &
  \multicolumn{2}{c|}{Ohsumed} &
  \multicolumn{2}{c|}{TREC} &
  \multicolumn{2}{c|}{DBLP} &
  \multicolumn{2}{c|}{Biomedical} &
  \multicolumn{2}{c}{CR} \\ \cmidrule{2-13} 
         & Acc   & F1    & Acc   & F1    & Acc   & F1    & Acc   & F1    & Acc   & F1    & Acc   & F1    \\ \hline
\textit{w/o SSHM} & 94.59 & 82.93 & 66.59 & 56.25 & 94.20 & 93.97 & 74.49 & 67.75 & 59.28 & 58.45 & 79.69 & 74.46 \\
\textit{w/o CL}   & 95.09 & 83.52 & 66.86 & 57.52 & 94.40 & 93.30 & 74.65 & 68.04 & 59.44 & 58.98 & 77.39 & 69.75 \\
\textit{w/o CLS}  & 95.62 & 83.95 & 66.89 & 57.56 & 94.42 & 93.99 & 74.69 & 68.22 & 59.66 & 60.02 & 78.59 & 72.16 \\
\textit{w/o SS} & 95.66 & 83.99 & 67.28 & 59.13 & 93.80 & 84.88 & 75.53 & 71.43 & 61.35 & 60.72 & 80.05 & 75.15 \\
\textit{w/o HM} & 95.69 & 84.05 & 67.55 & 59.95 & 94.00 & 93.99 & 75.44 & 70.78 & 60.92 & 60.52 & 82.78 & 80.18 \\
\textit{w/o MI} & 95.02 & 83.75 & 67.12 & 59.52 & 93.89 & 93.56 & 74.96 & 69.68 & 60.36 & 59.59 & 83.15 & 79.90  \\
\textit{w/o $\delta_{\{\text{CLS,CL}\}}$}    & 95.32 & 84.33 & 67.86 & 60.22 & 94.40 & 94.43 & 75.39 & 70.05 & 60.82 & 60.32 & 83.49 & 80.51 \\
\textit{w/o $\delta_{\text{CLS}}$}     & 95.79 & 84.59 & 68.19 & 60.46 & 94.46 & 94.55 & 75.52 & 70.26 & 60.93 & 60.46 & 83.52 & 80.68 \\
\textit{w/o $\delta_{\text{CL}}$} & 95.95 & 84.65 & 68.22 & 60.52 & 94.68 & 94.62 & 75.64 & 70.68 & 61.36 & 60.59 & 83.75 & 80.90  \\
Ours &
  \textbf{96.11} &
  \textbf{84.83} &
  \textbf{68.44} &
  \textbf{60.82} &
  \textbf{94.80} &
  \textbf{94.83} &
  \textbf{76.30} &
  \textbf{71.15} &
  \textbf{61.56} &
  \textbf{60.78} &
  \textbf{84.04} &
  \textbf{81.12} \\ \bottomrule 
\end{tabular}%
\end{table*}

\begin{figure*}[ht]
    \centering
    \subfigure[R52]{\includegraphics[width=0.21\textwidth]{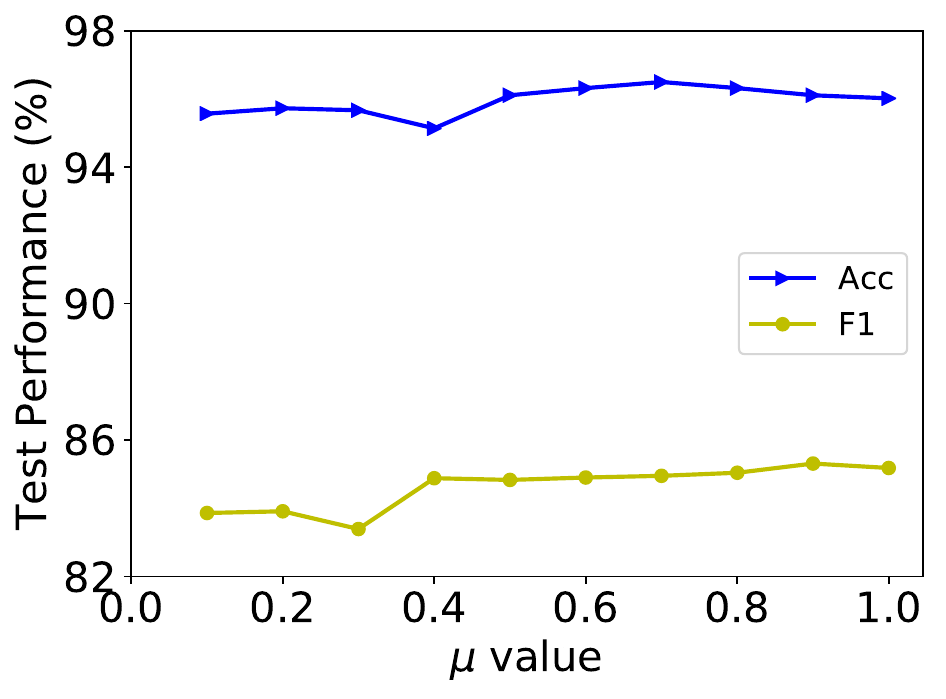}}
    \subfigure[R52]{\includegraphics[width=0.21\textwidth]{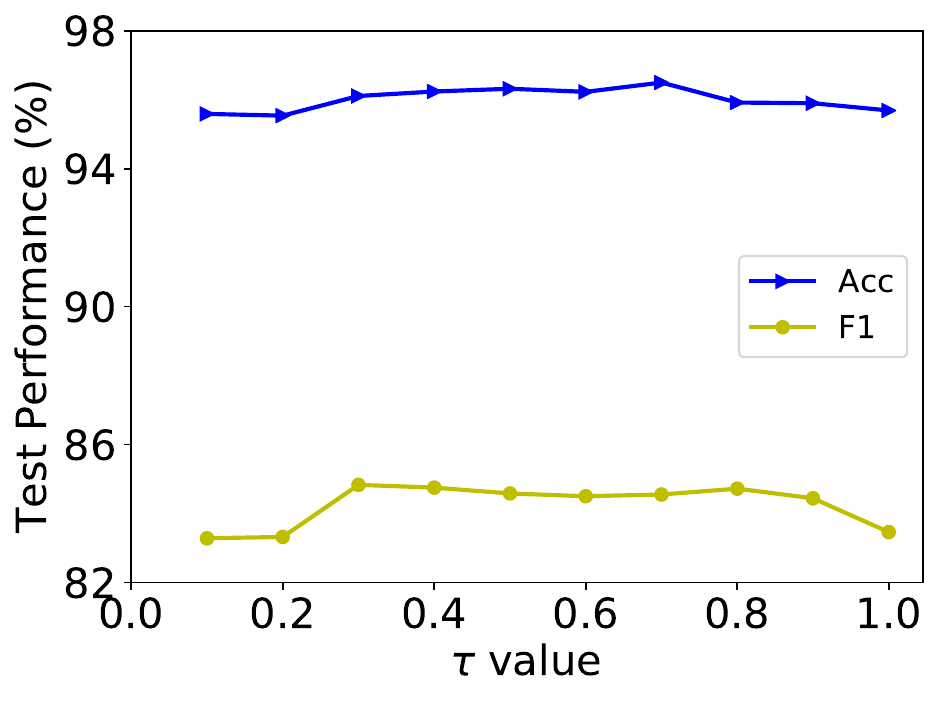}}
    \subfigure[Ohsumed]{\includegraphics[width=0.21\textwidth]{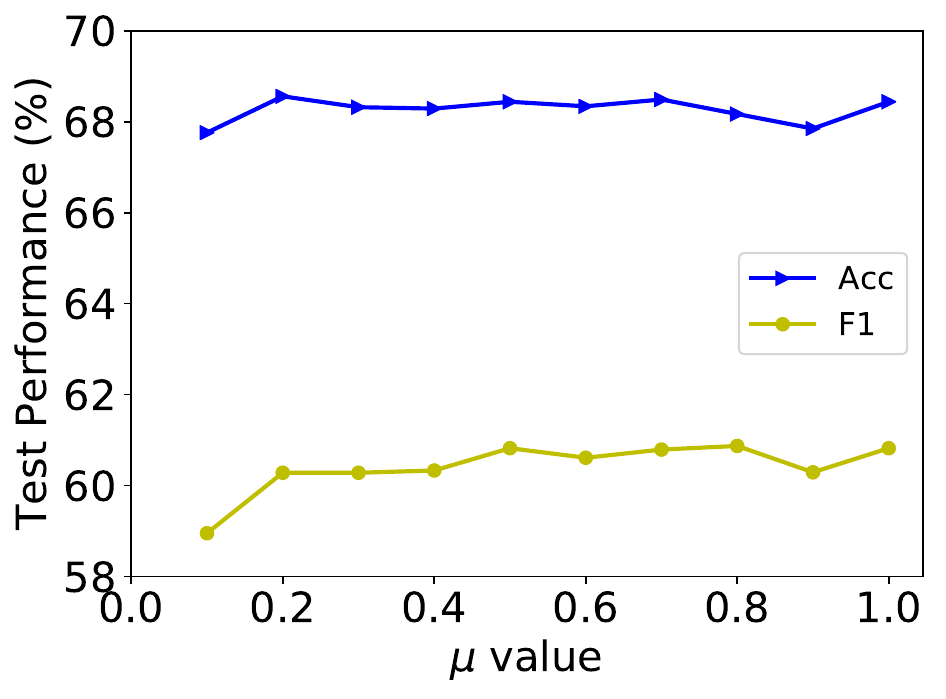}}
    \subfigure[Ohsumed]{\includegraphics[width=0.21\textwidth]{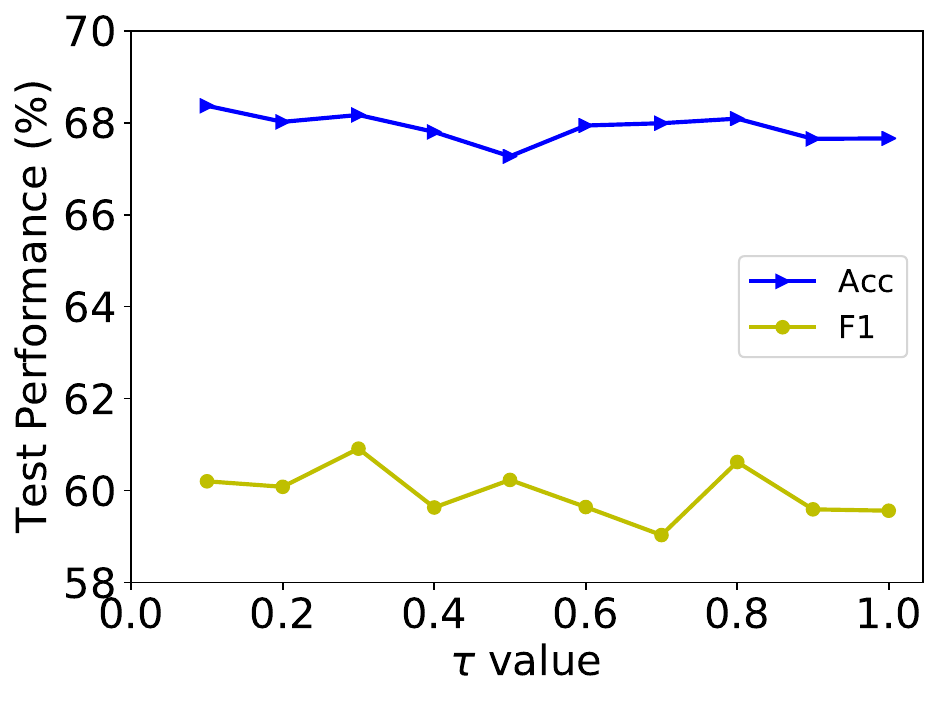}}
    \caption{Sensitivity of SharpReCL with respective to $\mu$ and $\tau$ on R52 and Ohsumed datasets.} 
    \label{hyper}
\end{figure*}


\noindent \textbf{Ablation Study.}
To evaluate the individual effects of the proposed SharpReCL, we perform a series of ablation experiments on all datasets except Ohsumed with an imbalance ratio of 50. We use the original Ohsumed dataset. Specifically, \textit{w/o SSHM} removes the simple-sampling and hard-mixup module, which simply incorporates the cross-entropy and CL objective. \textit{w/o SS} and \textit{w/o HM} delete the simple-sampling module and hard-mixup module, respectively. \textit{w/o CL} excludes the CL branch and leaves the classification branch. \textit{w/o CLS} eliminates the classification branch, which is evaluated with a linear classifier by frozen representations trained via CL. \textit{w/o MI} removes the mutual guidance between the classification branch and the SCL branch, allowing both branches to learn in parallel like other SCL models, simply summing their losses. \textit{w/o $\delta_{\{\text{CLS,CL}\}}$} removes the class priors $\delta$ from both the classification branch shown in Eq.\ref{lc} and the CL branch shown in Eq.\ref{CL}. \textit{w/o $\delta_{\text{CLS}}$} removes the class prior $\delta$ in the classification branch. \textit{w/o $\delta_{\text{CL}}$} eliminates the class prior $\delta$ in the CL branch. The results are presented in Table \ref{ablation}. 

We observe that each component of SharpReCL is indispensable and that removing any of the components degrades the performance. The first variant has the most significant performance degradation because the imbalanced class distribution in a batch makes the model learn biased text representations. The fourth and fifth variants show that simple-sampling and hard-mixup modules are critical for the quality of the rebalanced dataset. Removing the CL branch drastically reduces the ability of the model to extract invariant knowledge from similar texts. 

\noindent \textbf{Parameter Sensitivity.} 
We investigate the sensitivity of the model with respect to primary hyperparameters on the Ohsumed dataset, \textit{i.e.}, the controlled weight for contrastive loss $\mu$ and the temperature $\tau$. From Fig. \ref{hyper}, we find that the model is insensitive to the parameter $\mu$ from 0.2 to 1, illustrating its robustness. Moreover, the accuracy of the model is relatively stable across different selections of $\tau$ from [0.1, 1], while the F1 scores vary drastically depending on different $\tau$. 
We present other hyperparameters in \textbf{Appendix} \ref{hyper_appendix}.

\begin{figure}
    \centering
    \includegraphics[width=0.15\textwidth]{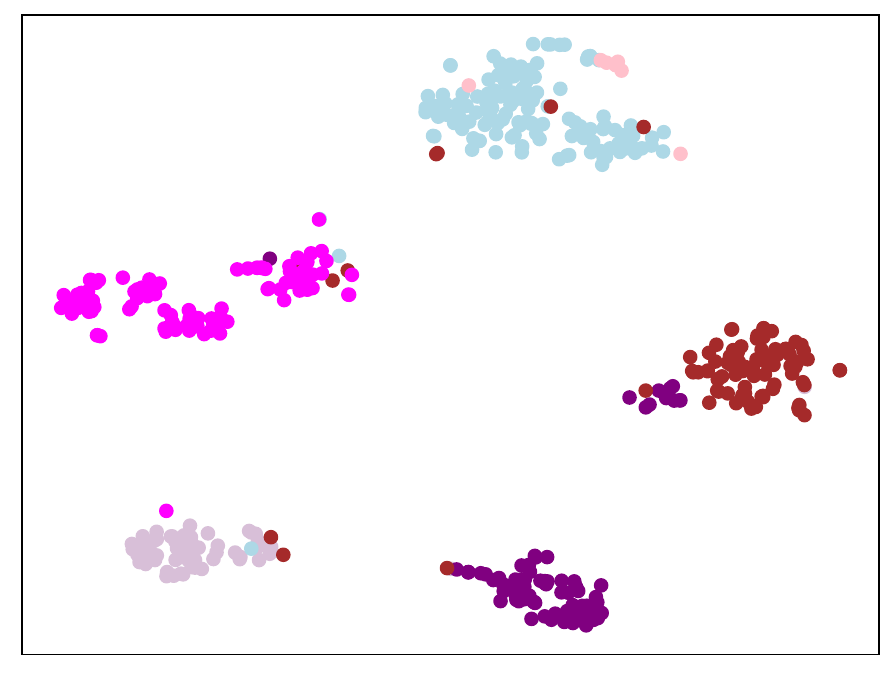}
    \includegraphics[width=0.15\textwidth]{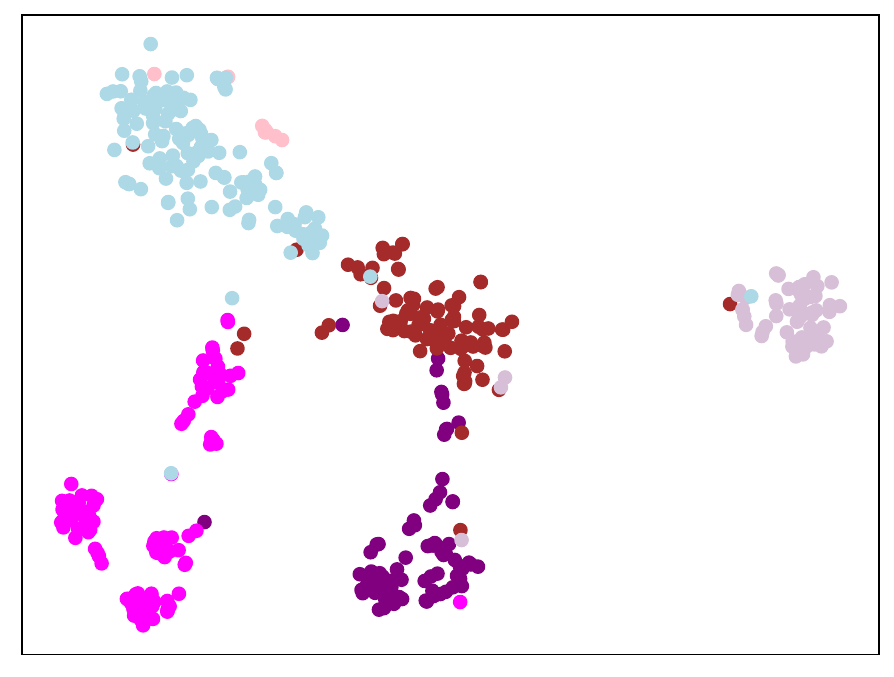}
    \includegraphics[width=0.15\textwidth]{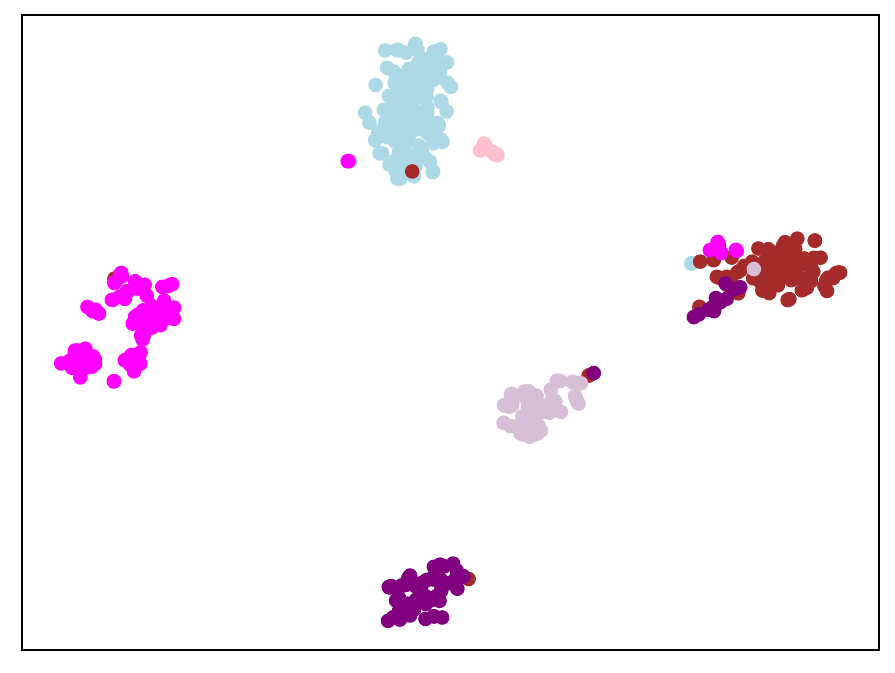}
    \caption{Visualization of different models on TREC under ir=50. left: SCLCE, middle: SPCL, right: Our method.}
    \label{visualize}
\end{figure}

\noindent \textbf{Visualization Study.} 
To show the effectiveness of our model in learning text representations under an imbalanced scenario, we perform visualizations of the embeddings of test documents by using the t-SNE method \cite{van2008visualizing}. We choose two representative models, SCLCE and SPCL, to compare with our model. From Fig. \ref{visualize}, we observe that the embeddings learned by SharpReCL are more discriminative and clustered than others.
\section{Conclusion}
In this work, we propose a novel model, namely, SharpReCL, for imbalanced TC tasks based on SCL. We make the classification branch and the SCL branch communicate and guide each other by leveraging learned prototype vectors.
To address the sensitivity of SCL to class imbalance, we construct a balanced dataset consisting of hard negative and hard positive samples by using the mixup technique to balance the sample pairs for SCL training. Extensive experiments on several datasets under different imbalanced settings demonstrate the effectiveness of our model.

\begin{acks}
Our work is supported by the National Natural Science Foundation of China (No. 62372209). Fausto Giunchiglia's work is funded by European Union's Horizon 2020 FET Proactive Project (No.823783).
\end{acks}
\bibliographystyle{ACM-Reference-Format}
\balance
\bibliography{sample-base}
\renewcommand{\thetable}{A.\arabic{table}}
\captionsetup[table]{labelformat=simple, labelsep=colon, name=Table}
\setcounter{table}{0}

\renewcommand{\thefigure}{A.\arabic{figure}}
\captionsetup[figure]{labelformat=simple, labelsep=colon, name=Figure}
\setcounter{figure}{0}
\appendix

\section{Appendix}
\subsection{The Impacts of Different Data Augmentations}
\label{data_aug}
To investigate the impacts of different data augmentations on the model, we implement the following three methods. (I) \textit{Word substitution}: We use synonyms from WordNet to replace associated words in the input text. (II) \textit{Back translation}: We first translate the input text into another language (French), and then translate it back into English to generate a paraphrased version of the input text. (III) \textit{Contextual embedding}: We use pre-trained LMs to find the top-n most suitable words in the input text for insertion. Based on the results shown in Table \ref{augmentation}, we can observe that all data augmentation methods can achieve desirable results. However, there is no universal method that can perform optimally on all datasets, as the optimal configuration of data augmentation methods varies depending on the dataset.

\begin{table*}[ht]
\caption{The results of different augmentations on all original datasets.}
\label{augmentation}
\centering
\begin{tabular}{c|cc|cc|cc|cc|cc|cc}
\hline
\multirow{2}{*}{Dataset} &
  \multicolumn{2}{c|}{R52} &
  \multicolumn{2}{c|}{Ohsumed} &
  \multicolumn{2}{c|}{TREC} &
  \multicolumn{2}{c|}{DBLP} &
  \multicolumn{2}{c|}{Biomedical} &
  \multicolumn{2}{c}{CR} \\ \cline{2-13} 
                     & Acc   & F1    & Acc   & F1    & Acc   & F1    & Acc   & F1    & Acc   & F1    & Acc   & F1    \\ \hline
Word substitution    & 96.11 & 84.83 & 68.44 & 60.82 & 97.20 & 97.63 & 80.15 & 77.59 & 72.52 & 72.65 & 93.09 & 92.51 \\
Back translation     & 95.92 & 84.69 & 68.09 & 59.41 & 97.20 & 96.84 & 80.12 & 77.22 & 72.83 & 72.87 & 93.09 & 92.58 \\
Contextual embedding & 96.05 & 84.76 & 67.62 & 59.20 & 97.80 & 97.18 & 80.14 & 77.38 & 72.61 & 72.61 & 93.09 & 92.54 \\ \hline
\end{tabular}%
\end{table*}

\subsection{Detailed Descriptions of Datasets}
\label{des_dataset}
\textbf{R52} \cite{liu2021deep} is a dataset containing news articles of 52 categories from Reuters for news classification. \textbf{Ohsumed} \cite{linmei2019heterogeneous} contains many important medical studies and a bibliographic classification dataset. Notably, this dataset contains only the titles. \textbf{TREC} \cite{li2002learning} is a question classification dataset that includes six categories of questions. \textbf{DBLP} \cite{tang2015pte} consists of six diverse kinds of paper titles extracted from computer science bibliographies. \textbf{Biomedical} \cite{xu2017self} is a collection of biomedical paper titles that is used for topic classification. \textbf{CR} \cite{ding2008holistic} is a customer review dataset for sentiment analysis, where the reviews are labeled positive or negative.

\subsection{Detailed Descriptions of Baselines}
\label{des_baseline}
\textbf{Pretrained language models and their imbalanced learning versions} 

(I) \textbf{BERT} \cite{devlin2018bert} is trained with a large-scale corpus and encodes implicit semantic information. (II) \textbf{BERT+Dice Loss} \cite{li2020dice} uses BERT to encode texts and the Dice loss for optimization. The Dice method dynamically adjusts the weights according to the learning difficulty of the input samples. (III) \textbf{BERT+Focal Loss} \cite{lin2017focal} combines BERT, which learns text features, with Focal loss, which offers small scaling factors for predicted classes with high confidence. (IV) \textbf{RoBERTa} \cite{liu2019roberta} is an enhanced version of BERT that utilizes more training data and large batch sizes. 

\noindent \textbf{Contrastive learning models}

(I) \textbf{SimCSE} \cite{gao2021simcse} uses two dropout operations in an unsupervised manner and sentence labels in a supervised manner to generate positive sample pairs. (II) \textbf{SCLCE} \cite{gunel2020supervised} combines the SCL objective with cross-entropy in the fine-tuning stage of NLP classification models. (III) \textbf{SPCL} \cite{song2022supervised} attempts to address the class-imbalanced task by combining CL and curriculum learning. (IV) \textbf{DualCL} \cite{chen2022dual} solves TC tasks via a dual CL mechanism, which simultaneously learns the input text features and the classifier parameters.

\noindent \textbf{Imbalanced contrastive learning models}

(I) \textbf{Hybrid-SC} \cite{wang2021contrastive} is a hybrid network architecture that integrates a supervised contrastive loss for image representation learning and a cross-entropy loss for classifier learning. The training process is designed to gradually shift from feature learning to classifier optimization. (II) \textbf{BCL} \cite{zhu2022balanced} simultaneously balances the gradient contributions of negative classes and ensures the presence of all classes within each mini-batch. This design satisfies the conditions for forming a regular simplex, which facilitates the optimization of the cross-entropy loss. (III) \textbf{TSC} \cite{li2022targeted} first generates a set of uniformly distributed target vectors on a hypersphere. During training, it encourages features from different classes to align with these distinct targets, thereby enforcing a uniform distribution of all classes in the feature space..

\noindent \textbf{Large language models} 

(I) \textbf{GPT-3.5} \cite{ouyang2022training} leverages a vast amount of texts for self-supervised learning and incorporates reinforcement learning with human feedback techniques to fine-tune the pretrained model. Note that when we conduct the experiments, the GPT-3.5 API is still available. (II) \textbf{Bloom-7.1B} \cite{workshop2022bloom} is a decoder-only Transformer language model trained on the ROOTS corpus, and exhibits enhanced performance after being fine-tuned with multi-task prompts. (III) \textbf{Llama2-7B} \cite{touvron2023llama} is also a decoder-only large language model that is trained on a new mixup of data from public dataset. It also increases the size of the pretraining corpus by 40\%, doubles the model's context length, and employs grouped query attention. (IV) \textbf{Llama3-8B} \cite{llama3modelcard} is also an auto-regressive language model, adopting a similar model architecture to Llama2. The main difference lies in its pretraining on a dataset containing over 15T tokens.

\subsection{Hyperparameter Study}
\label{hyper_appendix}
\begin{figure*}[ht]
    \centering
    \subfigure[TREC]{\includegraphics[width=0.23\textwidth]{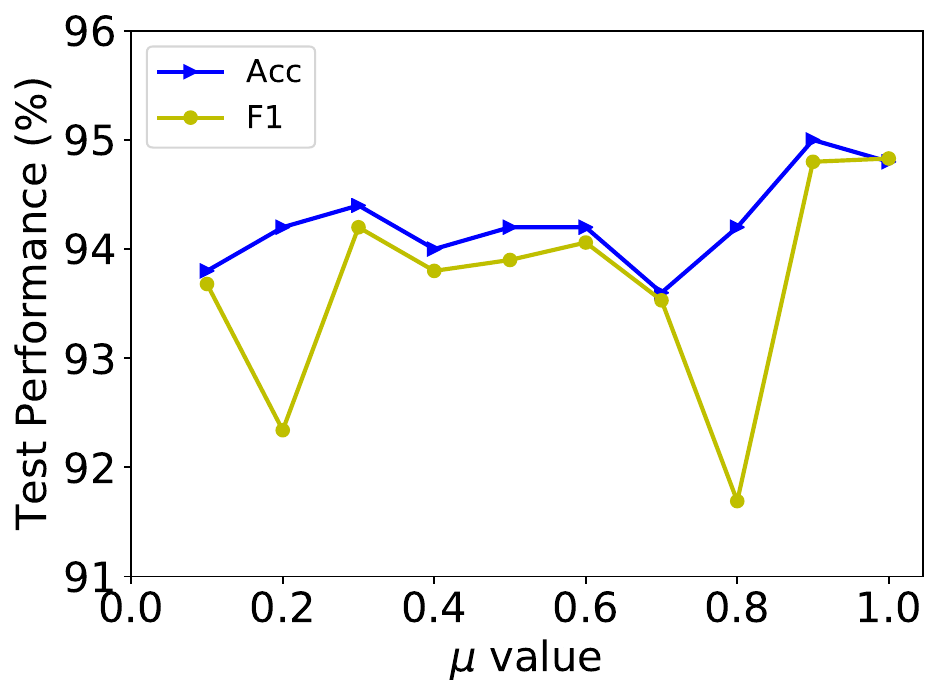}}
    \subfigure[TREC]{\includegraphics[width=0.23\textwidth]{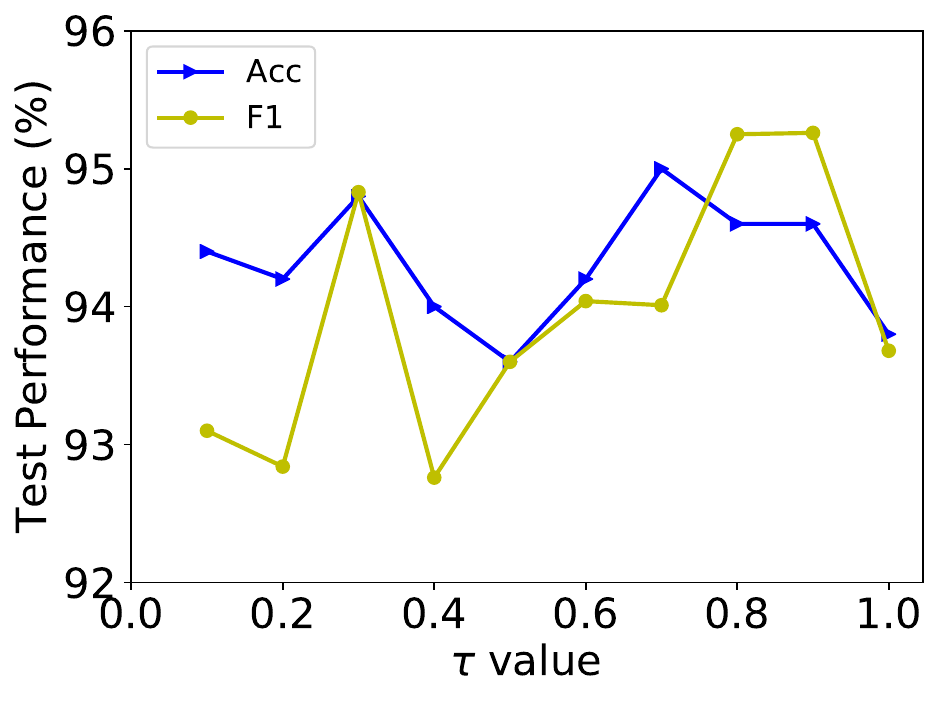}}
    \subfigure[DBLP]{\includegraphics[width=0.23\textwidth]{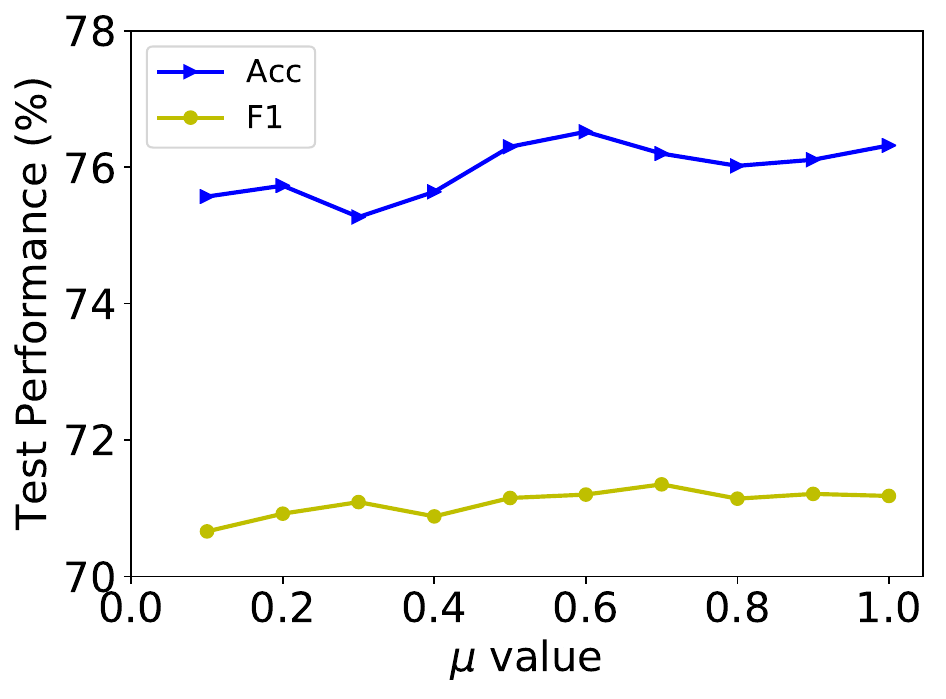}}
    \subfigure[DBLP]{\includegraphics[width=0.22\textwidth]{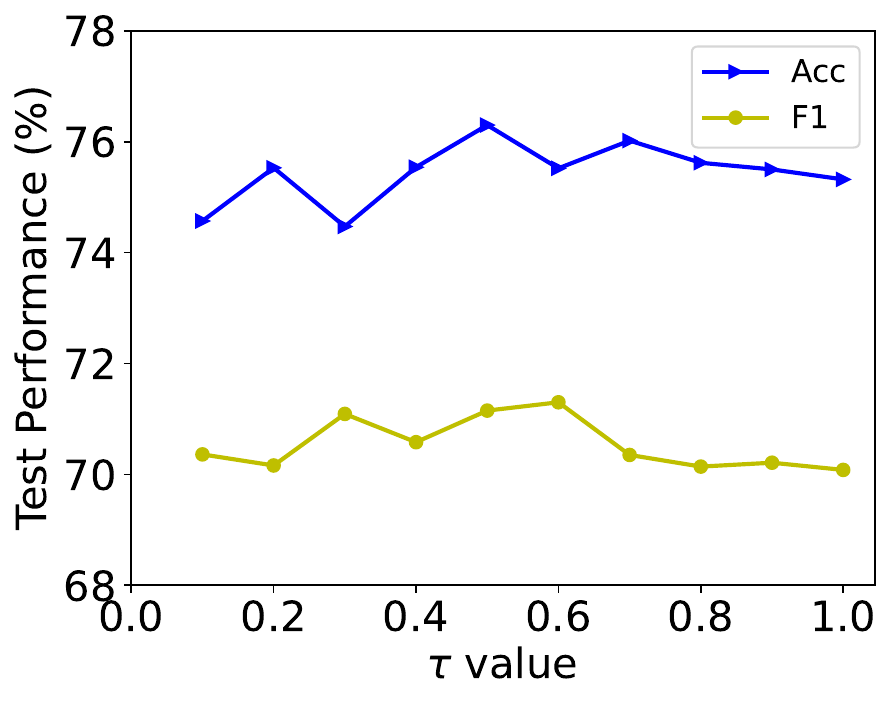}}
    \caption{Hyperparameter sensitivity of our model on TREC and DBLP under $ir$=50.}
    \label{trec_dblp}
\end{figure*}

\begin{figure*}[!ht]
    \centering
    \subfigure[Biomedical]{\includegraphics[width=0.23\textwidth]{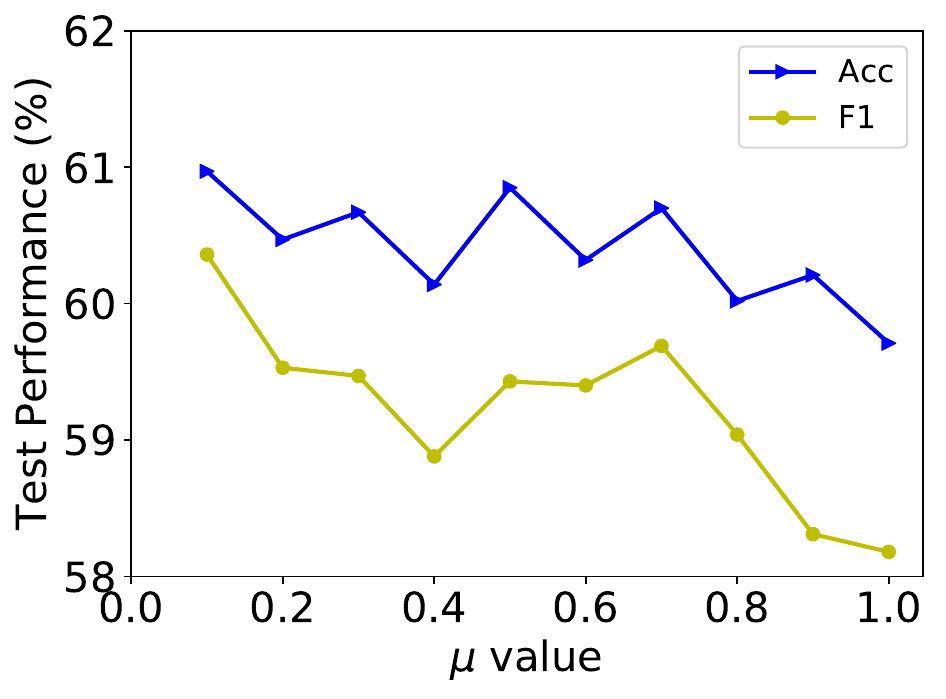}}
    \subfigure[Biomedical]{\includegraphics[width=0.23\textwidth]{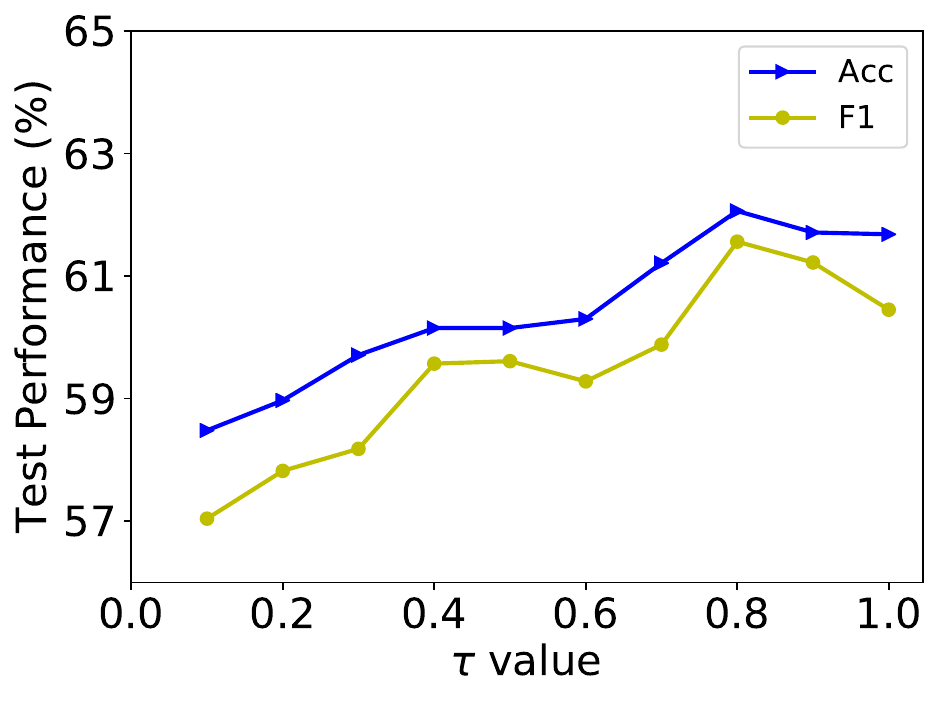}}
    \subfigure[CR]{\includegraphics[width=0.23\textwidth]{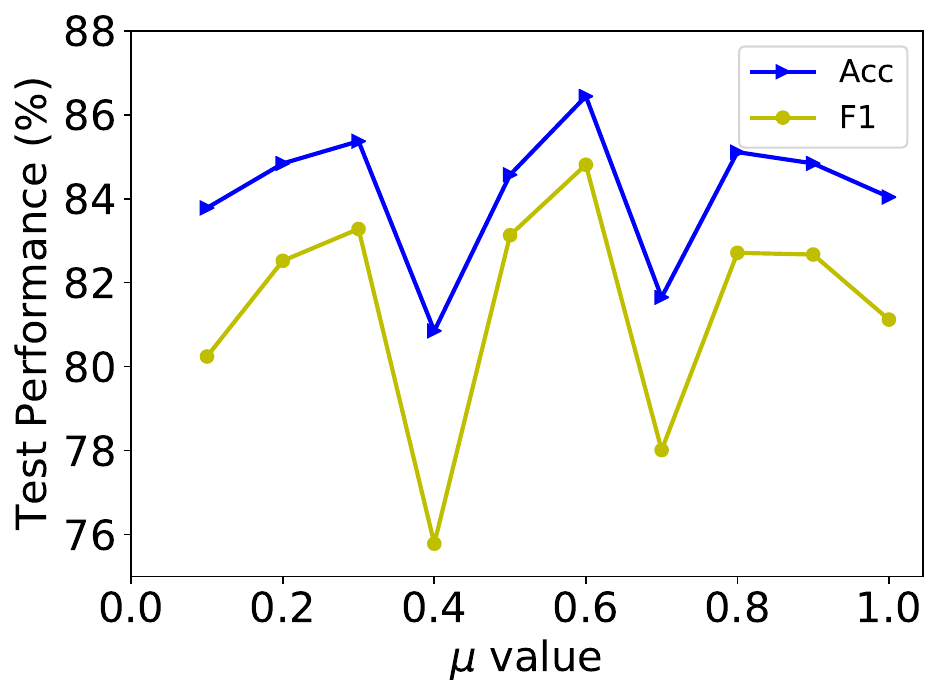}}
    \subfigure[CR]{\includegraphics[width=0.23\textwidth]{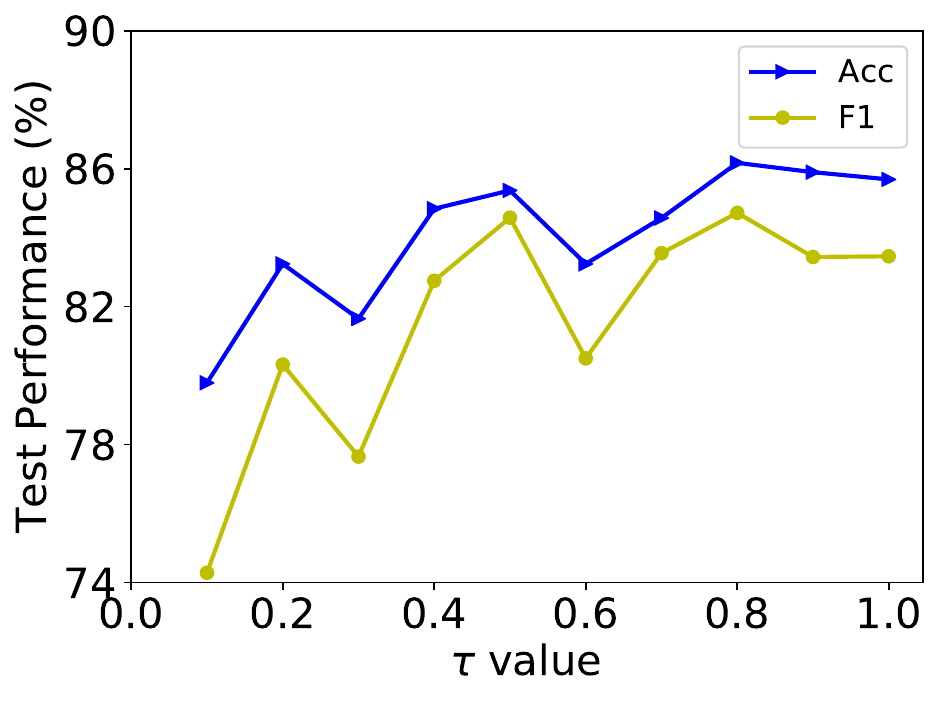}}
    \caption{Hyperparameter sensitivity of our model on Biomedical and CR under $ir$=50.}
    \label{Biomedical_CR}
\end{figure*}

\begin{table*}[!h]
\centering
\caption{Results of SharpReCL on the evaluated datasets.}
\label{k_value}
\begin{tabular}{c|ccccc}
\hline
Model & K=10       & K=20       & K=30       & K=40       & K=50       \\ \hline
R52                & 95.82$\pm$0.22 & \textbf{96.11$\pm$0.36} & 96.01$\pm$0.25 & 95.08$\pm$0.29 & 95.32$\pm$0.10 \\ \hline
Ohsumed                & 68.02$\pm$0.19 & \textbf{68.44$\pm$0.22} & 68.19$\pm$0.20 & 67.68$\pm$0.35 & 67.32$\pm$0.16 \\ \hline
TREC                & 94.02$\pm$0.49 & \textbf{94.80$\pm$0.55} & 94.29$\pm$0.26 & 93.68$\pm$0.32 & 93.36$\pm$0.12 \\ \hline
DBLP                & 75.22$\pm$0.29 & 76.30$\pm$0.25 & \textbf{76.59$\pm$0.22} & 76.08$\pm$0.39 & 75.92$\pm$0.26 \\ \hline
Biomedical                & 61.02$\pm$0.19 & \textbf{61.56$\pm$0.28} & 61.09$\pm$0.20 & 60.68$\pm$0.45 & 60.52$\pm$0.16 \\ \hline
CR                & 83.62$\pm$0.15 & \textbf{84.04$\pm$0.25} & 83.74$\pm$0.19 & 83.02$\pm$0.20 & 82.64$\pm$0.22 \\ \hline
\end{tabular}%
\end{table*}

We conduct a series of hyperparameter sensitivity studies with controlled weight $\mu$ and temperature $\tau$ on different datasets under \textit{ir}=50. From Figs. \ref{trec_dblp} and \ref{Biomedical_CR}, we observe that the best configuration of the $\mu$ and $\tau$ values varies with the different datasets. When the range of $\mu$ is [0.3, 0.7], this often leads to improved performance. For $\tau$, taking a value from [0.5, 0.9] allows the model to achieve satisfactory performance.

We also show the impact of the K value in the top-K positive samples on the evaluated datasets in the Table \ref{k_value}, where except for R52 and Ohsumed, the remaining datasets are conducted under the setting of $ir\!=\!50$. We observe that the model performance first increases and then decreases as K increases. The model performance can basically reach its peak at K=20. One reasonable reason is that when the value of K is too small, the model cannot synthesize diverse hard samples, while when the value of K is too large, the model may generate simple samples.

\end{document}